\documentclass[runningheads]{llncs}
\usepackage{graphicx}

\usepackage{tikz}
\usepackage{makecell}
\usepackage{comment}
\usepackage{amsmath,amssymb} %
\usepackage{booktabs}
\usepackage{float}
\usepackage{adjustbox}
\usepackage{multirow}
\usepackage{diagbox}
\usepackage{breakcites}
\usepackage{gensymb}
\usepackage{wrapfig}
\usepackage[T1]{fontenc}
\usepackage[utf8]{inputenc}
\usepackage[accsupp]{axessibility}  %

\DeclareUnicodeCharacter{FFFC}{ }
\newif\ifarxiv
\arxivtrue
\newcommand{\arxivoreccv}[2]{\ifarxiv #1\else #2\fi}
\usepackage{color,xcolor}
\usepackage[colorlinks]{hyperref}
\begin{document}
\pagestyle{headings}
\mainmatter
\def\ECCVSubNumber{2863}  %

\title{BlobGAN: Spatially Disentangled \\ Scene Representations} %

\titlerunning{BlobGAN: Spatially Disentangled Scene Representations}
\author{Dave Epstein\textsuperscript{\tiny 1} \qquad  Taesung Park\textsuperscript{\tiny 2} \qquad  Richard Zhang\textsuperscript{\tiny 2} \\
Eli Shechtman\textsuperscript{\tiny 2} \qquad  Alexei A. Efros\textsuperscript{\tiny 1}}
\authorrunning{D. Epstein, T. Park, R. Zhang, E. Shechtman, A.A. Efros}
\institute{\textsuperscript{\tiny 1}UC Berkeley \quad \textsuperscript{\tiny 2}Adobe Research}
\maketitle
\begin{abstract}
We propose an unsupervised, mid-level representation for a generative model of scenes. The representation is mid-level in that it is neither per-pixel nor per-image; rather, scenes are modeled as a collection of spatial, depth-ordered ``blobs'' of features. 
Blobs are differentiably placed onto a feature grid that is decoded into an image by a generative adversarial network. 
Due to the spatial uniformity of blobs and the locality inherent to convolution, our network learns to associate different blobs with different entities in a scene and to arrange these blobs to capture scene layout.
We demonstrate this emergent behavior by showing that, despite training without any supervision, our method enables applications such as easy manipulation of objects within a scene ({\em e.g.} moving, removing, and restyling furniture), creation of feasible scenes given constraints ({\em e.g.} plausible rooms with drawers at a particular location), and parsing of real-world images into constituent parts. On a challenging multi-category dataset of indoor scenes, BlobGAN outperforms StyleGAN2 in image quality as measured by FID. See our project page for video results and interactive demo:  \href{http://www.dave.ml/blobgan}{http://www.dave.ml/blobgan}. 
\end{abstract}

\newcommand\blfootnote[1]{%
  \begingroup
  \renewcommand\thefootnote{}\footnote{#1}%
  \addtocounter{footnote}{-1}%
  \endgroup
}

\begin{figure}[t]
\centering
\includegraphics[width=0.8\textwidth]{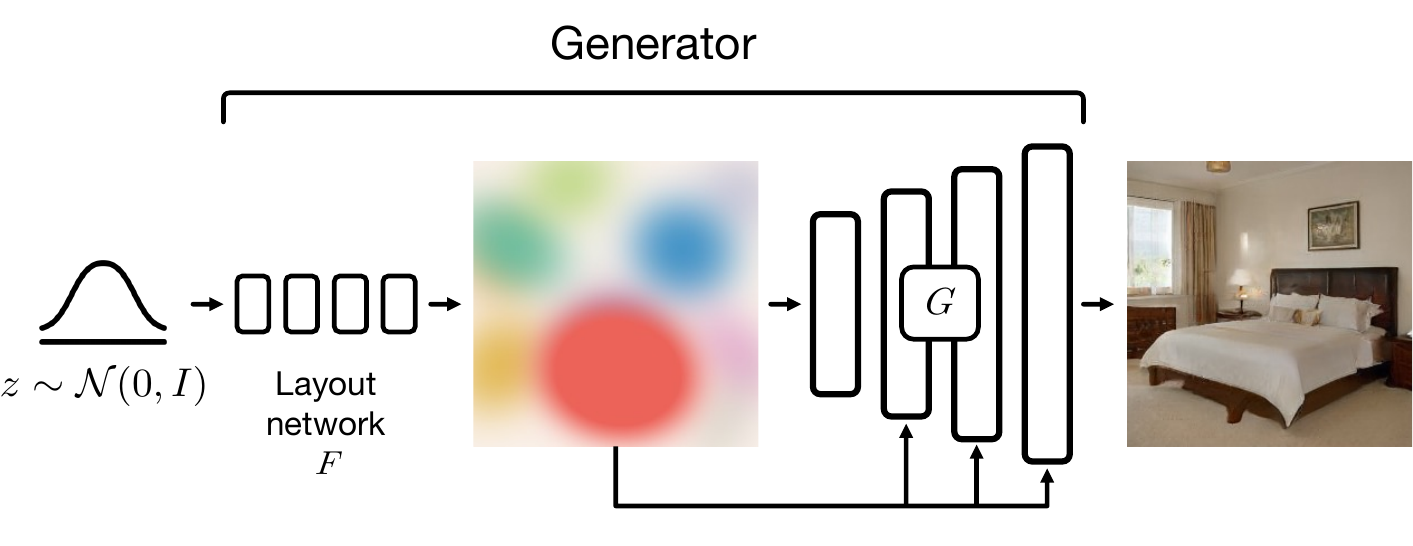}
\vspace{-2mm}
\caption{In our generator, random noise is mapped by the layout network $F$ to blob parameters.
 Blobs output by $F$ are \textit{splatted} spatially onto corresponding locations in the feature grid, used both as initial input and as spatially-adaptive modulation for the convolutional decoder $G$. Our blob representation automatically serves as a strong mid-level generative representation for scenes, discovering objects and their layouts.}
\label{fig:architecture}
\end{figure}

\section{Introduction}

The visual world is incredibly rich.  It is so much more than the typical ImageNet-style photos of solitary, centered objects (cars, cats, birds, faces, {\em etc.}), which are the mainstays of most current paper result sections.  Indeed, it was long clear, both in human vision~\cite{biederman1981semantics,hock1978real} and in computer vision~\cite{YakimovskyF73,Ohta-1978,oliva2007role,hoiem2007ijcv,gupta2010blocks}, that understanding and modeling objects within the context of a {\em scene} is of the utmost importance. Visual artists have understood this for centuries, first by discovering and following the rules of scene formation during the Renaissance, and then by expertly breaking such rules in the 20th century (cf. the surrealists including Magritte, Ernst, and Dal\'{i}).

However, in the current deep learning era, scene modeling for both analysis and synthesis tasks has largely taken a back seat.  Images of scenes are either represented in a top-down fashion, no different from objects -- {\em i.e.} for GANs or image classifiers, scene classes such as ``bedrooms" or ``kitchens" are represented the same way as object classes, such as ``beds" or ``chairs". Or, scenes are modeled in a bottom-up way by semantic labeling of each image pixel, {\em e.g.}, semantic segmentation, pix2pix~\cite{isola2017image}, SPADE~\cite{park2019semantic}, {\em etc}. 
Both paths seem unsatisfactory because neither can provide easy ways of reasoning about parts of the scene as {\em entities}.  The scene parts are either baked into a single entangled latent vector (top-down), or need to be grouped together from individual pixel labels (bottom-up).

In this paper, we propose an {\em unsupervised mid-level representation} for a generative model of scenes. 
The representation is mid-level in that it is neither
per-pixel nor per-image; rather, scenes are modeled as a collection of spatial, depth-ordered Gaussian ``blobs''.  
This collection of blobs provides a bottleneck in the generative architecture, as shown in Figure~\ref{fig:architecture}, forcing each blob to correspond to a specific object in the scene and thus causing a spatially disentangled representation to emerge. This representation allows us to perform a number of scene editing tasks (see Figure~\ref{fig:sequenceedits}) previously only achievable with extensive semantic supervision, if at all. %

\section{Related Work}

\noindent\textbf{Mid-level scene representations.} 
Work on mid-level scene representations can be traced back to the 1970s, to the seminal papers of Yakimovsky and Feldman~\cite{YakimovskyF73} and Ohta et al~\cite{Ohta-1978}, which already contained many key ideas including joint bottom-up segmentation and top-down reasoning.  Other important developments were the line of work on normalized-cuts segmentation~\cite{shi2000normalized,yu2002concurrent,hariharan2014simultaneous} and qualitative 3D scene interpretation~\cite{hoiem2007ijcv,hedau2009recovering,silberman2012indoor,gupta2010blocks} in the early 2000s. But most relevant to the present manuscript is the classic {\em Blobworld} work of Carson et al.~\cite{carson1999blobworld}, a region-based image retrieval system, with each image represented by a mixture-of-Gaussian blobs. Our model could be considered a generative version of this representation, except we also encode the depth ordering of the blobs. %

\noindent\textbf{Scene analysis by synthesis.} The idea of modeling a complex visual scene by trying to generate it has been attempted a number of times in the past. Early methods, such as \cite{tu2005image,sudderth2005learning,torralba2005describing}, introduced key ideas but were limited by the generative models of the time. To address this, several approaches tried non-parametric generation~\cite{malisiewicz2009beyond,russell2009segmenting,SceneCollaging}, with Scene Collaging~\cite{SceneCollaging} the most valiant attempt, showing layered scene representations despite very heavy computational burden. With the advancement of deep generative models, parametric analysis-by-synthesis techniques are having a renaissance, with some top-down~\cite{yao20183d,oktay2018counterfactual,yu2021unsupervised} as well as bottom-up~\cite{pathakCVPR16context,he2021masked} techniques.

\noindent\textbf{Conditional image generation.} 
Conditional GANs~\cite{zhu2017toward,huang2018multimodal,wang2018high}, such as image-to-image translation setups~\cite{isola2017image}, predict an image from a predefined representation, {\em e.g.} semantic segmentation maps~\cite{park2019semantic,li2021collaging}, object-attribute graphs ~\cite{johnson2018image,bear2020learning}, text~\cite{zhang2017stackgan,reed2016generative,glide,dalle2,rombach2021high,dalle}, pose~\cite{lewis2021tryongan,sarkar2021style,albahar2021pose}, and keypoints~\cite{brooks2021hallucinating}. Other setups include using perceptual losses~\cite{chen2017photographic}, implicit likelihood estimation~\cite{li2019diverse}, and more recently, diffusion models~\cite{meng2021sdedit,saharia2021palette}. \cite{siarohin2021motion,mejjati2020objectstamps,mejjati2021gaussigan,Siarohin_2019_NeurIPS,wang2022improving,he2021latentkeypointgan} explore related intermediate representations to help generation (mostly of humans or objects) but none provide the ability to generate and manipulate high-quality scene images of our method.

\noindent\textbf{Unconditional generation and disentanglement.} Rather than use explicit conditioning, it is possible to learn an image ``manifold'' with a generative model such as a VAE~\cite{kingma2013auto,higgins2016beta} or GAN~\cite{goodfellow2014generative} and explore emergent capabilities. GANs have improved in image quality~\cite{radford2015unsupervised,denton2015deep,zhang2019self,brock2018large,karras2017progressive,karras2019style,sgan2,karras2021alias} and are our focus. Directions of variation naturally emerge in the latent space and can be discovered when guided by geometry/color changes~\cite{jahanian2019steerability}, language or attributes~\cite{patashnik2021styleclip,radford2015unsupervised,shen2020interfacegan,abdal2021styleflow,wu2021stylespace}, cognitive signals~\cite{goetschalckx2019ganalyze}, or in an unsupervised manner~\cite{harkonen2020ganspace,shen2021closed,peebles2020hessian}. Discovering disentangled representations remains a challenging open problem~\cite{locatello2019challenging}. To date, most successful applications have been on data of objects, {\em e.g.} faces and cars, or changing textures for scenes~\cite{park2020swapping}. Similar to us, an active line of work explores adding 3D inductive biases~\cite{Niemeyer2020GIRAFFE,nguyen2019hologan,nguyen2020blockgan}, but individual object manipulation has largely focused on simple diagnostic scenes~\cite{johnson2017clevr}.
Alternatively, the internal units of a pretrained GAN offer finer spatial control, with certain units naturally correlating with object classes~\cite{bau2018gan,bau2020rewriting,yang2021semantic}. The internal compositionality of GANs can be leveraged to harmonize images \cite{collins2020editing,chai2021using} or perform a limited set of edits on objects in a scene~\cite{bau2018gan,zhang2021decorating,zhu2022region}. Crucially, while these works require semantic supervision to identify units and regions, our work uses a representation where these factors naturally emerge.

\section{Method}

Our method aims to learn a representation of scenes as spatial maps of blobs through the generative process. As shown in Figure~\ref{fig:architecture}, a layout network maps from random noise to a set of blob parameters. Then, blobs are differentiably splatted onto a spatial grid -- a ``blob map'' -- which a StyleGAN2-like decoder~\cite{karras2019analyzing} converts into an image. Finally, the blob map is used to modulate the decoder We train our model in an adversarial framework with an unmodified discriminator \cite{sgan2}. Interestingly, even without explicit labels, our model learns to decompose scenes into entities and their layouts.

\begin{wrapfigure}{br}{0.\arxivoreccv{45}{45}\textwidth}
    \vspace{-18mm}
  \begin{center}
    \includegraphics[width=.\arxivoreccv{45}{45}\textwidth]{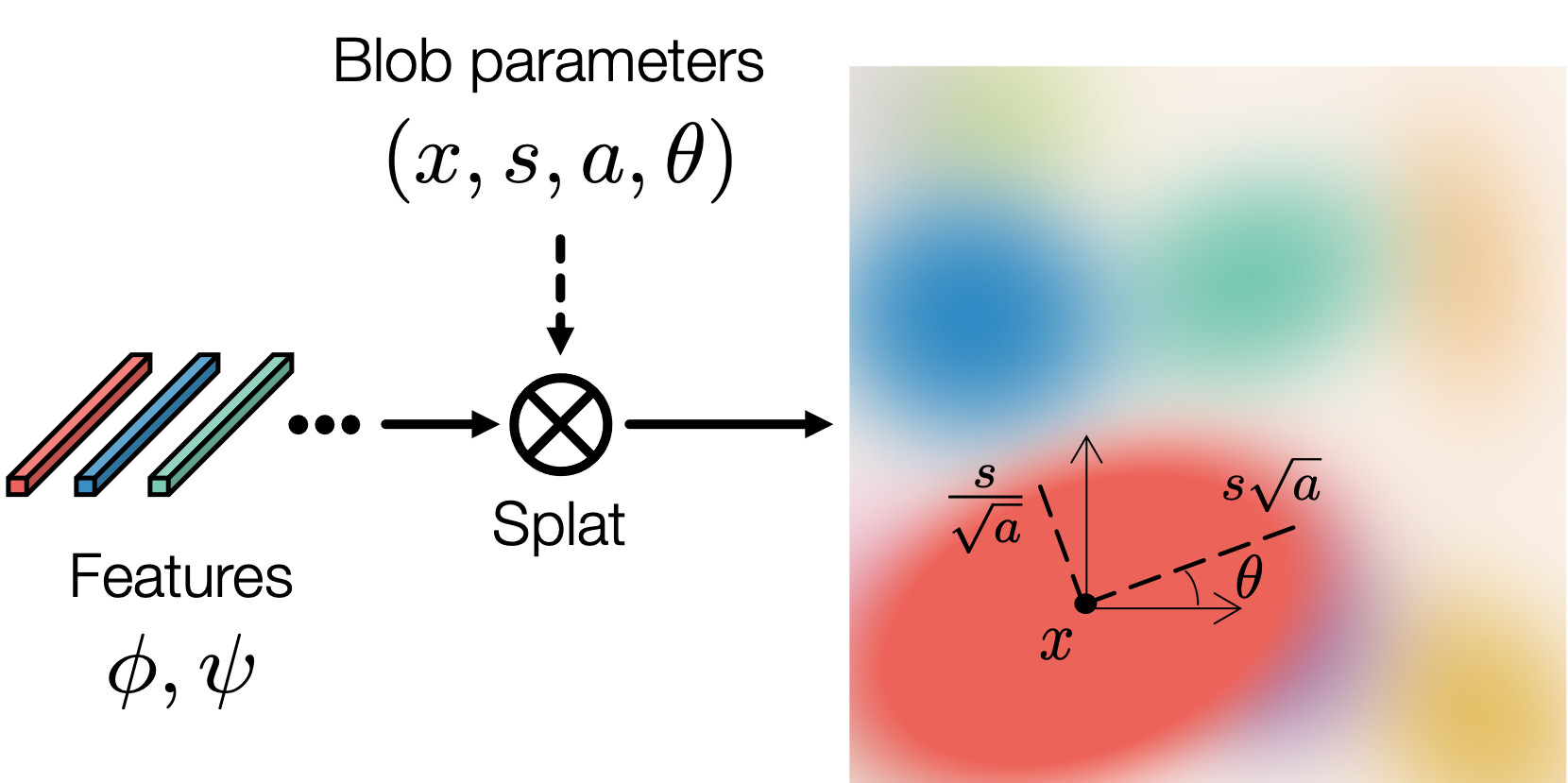}
  \end{center}
  \vspace{-3mm}
  \caption{\small Our elliptical blobs $\beta$ are parametrized by centroid $x$, scale $s$, aspect ratio $a$, and angle $\theta$. We composite multiple blobs with alpha values that smoothly decay from blob centers. The features $\phi$ or $\psi$ are splatted on their corresponding blobs and passed to the decoder.}
  \arxivoreccv{\vspace{-5mm}}{\vspace{-10mm}}
  \label{fig:blob_param}
\end{wrapfigure}

Our generator model is largely divided into two parts. First, we apply an 8-layer MLP $F$ to map random noise $z \in \mathbb{R}^{d_\text{noise}} \sim \mathcal{N}(0, I_d)$ to a collection of blobs parameterized by $\boldsymbol\beta = \{\beta_i\}_{i=1}^k$ which are splatted onto a spatial $H \times W \times d$ feature grid in a differentiable manner. This process is visualized in Figure~\ref{fig:blob_param}. The feature grid is then passed to a convolutional decoder $G$ to produce final output images. In the remainder of this section, we describe the design of our representation as well as its implementation in detail.

\subsection{From noise to blobs as layout}

\label{methods:blobs}
We map from random Gaussian noise to distributions of blobs with an MLP $F$ with dimension $d_\text{hidden}$. The last layer of $F$ is decoded into a sequence of blob properties $\boldsymbol\beta$. We opt for a simple yet effective parametrization of blobs, representing them as ellipses by their center coordinates $x \in [0,1]^2$, scale $s \in \mathbb{R}$, aspect ratio $a \in \mathbb{R}$, and rotation angle $\theta \in [-\pi,\pi]$. Each blob is also associated with a structure feature $\phi \in \mathbb{R}^{d_\text{in}}$ and a style feature $\psi \in \mathbb{R}^{d_\text{style}}$. Altogether, our blob representation is:
$$\beta \in \mathbb{R}^{2 + 1 + 1 + 1 + d_\text{in} + d_\text{style}} \triangleq \left(x, s, a, \theta, \phi, \psi \right)$$

Next, we transform the blob parameters to a 2D feature grid by populating the ellipse specified by $\beta$ with the feature vectors $\phi$ and $\psi$. We do this differentiably by assigning an opacity and spatial falloff to each blob. Specifically, we calculate a grid $\boldsymbol \alpha \in [0,1]^{H \times W \times k}$ which indicates each blob's opacity value at each location. We then use these opacity maps to \textit{splat} the features $\phi, \psi$ at various resolutions, using a single broadcasted matrix multiplication operation.

In more detail, we begin by computing per-blob opacity maps $o \in [0,1]^{H \times W}$. For each grid location $x_\text{grid} \in \left\{\left(\frac{w}{W},\frac{h}{H}\right)\right\}_{w=1, h=1}^{W, H}$ we find the squared Mahalanobis distance to the blob center $x$:
\begin{equation}
d(x_\text{grid}, x) = (x_\text{grid} - x)^T (R \Sigma R^T)^{-1} (x_\text{grid} - x),
\end{equation}
\noindent where $\Sigma = c \begin{bmatrix}
{a} & 0 \\
0 & \frac{1}{{a}} 
\end{bmatrix}$, $R$ is a 2D rotation matrix by angle $\theta$, and $c = 0.02$ controls blob edge sharpness.
The opacity of a blob at a given grid location is then:
\begin{equation}
o(x_\text{grid}) = \sigma\left(s - d(x_\text{grid}, x) \right),
\end{equation}
where $s$ acts as a control of blob size by shifting inputs to the sigmoid. Intuitively, this can be thought of as taking a soft thresholding operation on a Gaussian to define an in-region and an out-region. For example, our model can output a large negative $s < 0$ to effectively ``turn off'' a blob. Rather than taking the softmax to normalize values at each location, we use the alpha-compositing operation \cite{alphacomp}, which allows us to model occlusions and object relationships more naturally by imposing a 2.1-D z-ordering \cite{nitzberg19902}: 
\begin{equation}
    \alpha_i(x_\text{grid}) = o_i(x_\text{grid}) \prod_{j=i+1}^k (1 - o_j(x_\text{grid})).
\end{equation}
Lastly, our blob mapping network also outputs background vectors $\phi_0, \psi_0$, with a fixed opacity $o_0 = 1$. Final features at each grid location are the convex combination of blob feature vectors, given by the (k+1) $\alpha_i$ scores.

\subsection{From blob layouts to scene images}
\label{sec:blobtoimg}
We now describe a function $G$ that converts the representation of scenes as blobs $\boldsymbol\beta$ described in Section \ref{methods:blobs} into realistic, harmonized images. To do so, we build on the architecture of StyleGAN2 \cite{sgan2}. We modify it to take in a spatially-varying input tensor based on blob structure features rather than a single, global vector, and perform spatially-varying style modulation. 

As opposed to standard StyleGAN, where the single style vector $w$ must capture information about all aspects of the scene, our representation separates layout (blob locations and sizes) and appearance (per-blob feature vectors) by construction, naturally providing a foundation for a disentangled representation.

Concretely, we compute a feature grid $\Phi$ at $16 \times 16$ resolution using blob structure vectors $\phi_i$ 
and use $\Phi$ as input to $G$, removing the first two convolutional blocks of the base architecture to accommodate the increased resolution. We also apply spatial style-based modulation~\cite{park2019semantic} at each convolution using feature grids $\Psi_{l \times l} $ for $l \in \{16, 32, \ldots, 256\}$ computed from blob style vectors $\psi_i$.

\subsection{Encouraging disentanglement}
Intuitively, all activations within a blob are governed by the same feature vector, encouraging blobs to yield image regions of self-similar properties, {\em i.e.} entities in a scene. Further, due to the locality of convolution, the layout of blobs in the input must strongly inform the final arrangement of image regions. Finally, our latent space separates layout (blob location, shape, and size) from appearance (blob features) by construction. All the above help our model learn to bind individual blobs to different objects and arrange these blobs into a coherent layout, disentangling scenes spatially into their component parts.

To further nudge our network in this direction, we stochastically perturb blob representations $\boldsymbol\beta$ before inputting them to $G$, enforcing our model to be robust under these perturbations. We implement this by corrupting blob parameters with uniform noise $\delta x$, $\delta s$, and $\delta \theta$.
This requires that blobs be independent of each other, promoting object discovery and discouraging degenerate solutions which rely on precise blob placement or shape. 

We also experiment with style mixing, where with probability $0.2$ we uniformly sample between $0$ and $k$ blobs to swap, and permute style vectors for these blobs among different batch samples. We find that this intervention harms our training process since it requires that all styles match all layouts, an assumption we show does not hold in Section \ref{sec:layoutresults}. We also try randomly removing blobs from the forward process with some probability, but found this hurts training, since certain objects must always be present in certain kinds of scenes ({\em e.g.} kitchens are unlikely to have no refrigerator). This constraint led to a more distributed, and therefore less controllable, representation of scene entities.

\section{Experiments}
\label{sec:results}

\begin{figure}[t]
\centering
\includegraphics[width=\textwidth]{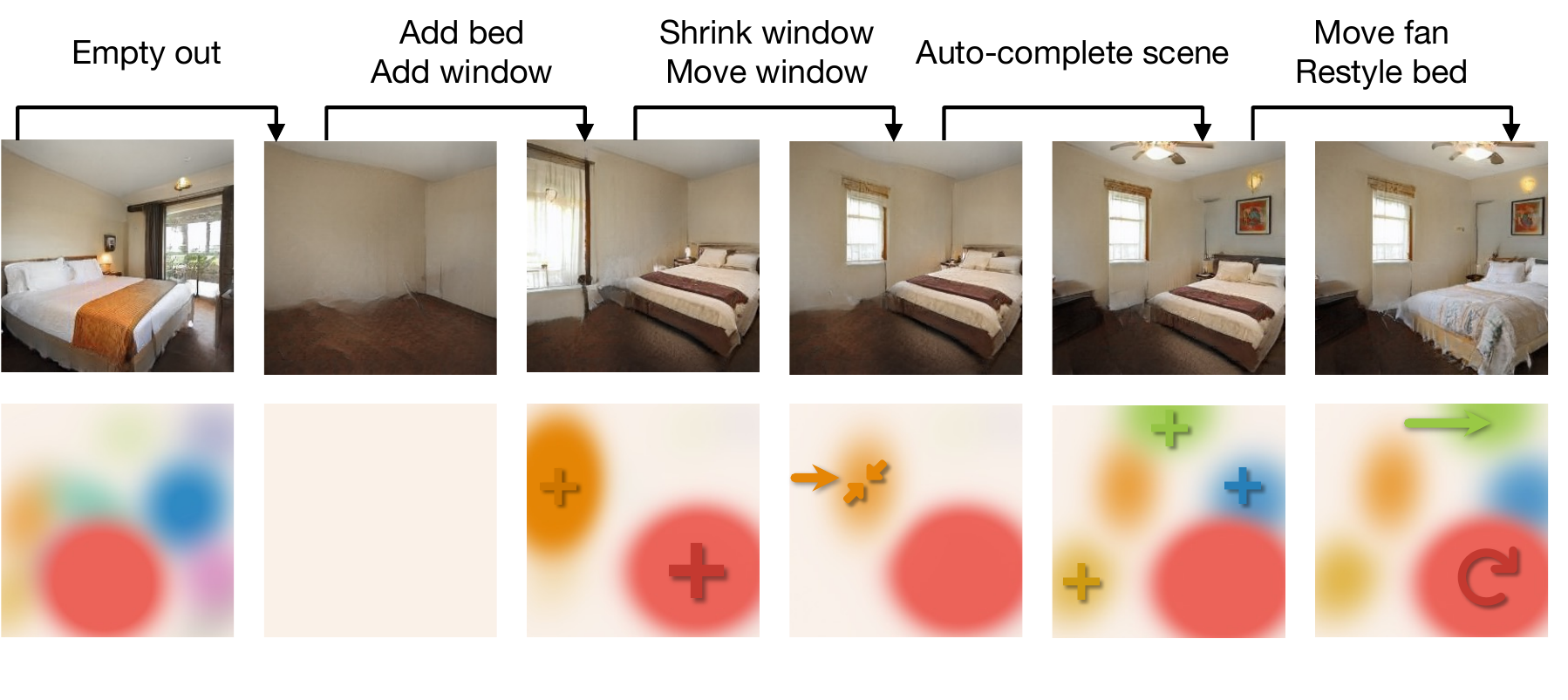}
\vspace{-.3in}
\caption{\textbf{Blobs allow extensive image manipulation:} We apply a sequence of modifications to the blob map of an image generated by our model and show the resulting images outputs at each stage of the editing process, demonstrating the strength of our learned representation.}
 \label{fig:sequenceedits}
\end{figure}

We evaluate our learned representation quantitatively and qualitatively and demonstrate that a spatially disentangled representation of scenes emerges. We begin by showing that our model learns to associate individual blobs with objects in scenes, and then show that our representation captures the distribution of scene layouts. We highlight some applications of our model in Figure \ref{fig:sequenceedits}. Finally, we use our model to parse the layouts of real scene images via inversion. For more results, including on additional datasets and ablations, please see Appendix.

\subsection{Training and implementation}
We largely follow the training procedure set forth in StyleGAN2 \cite{sgan2}, with nonsaturating loss \cite{goodfellow2014generative}, R1 regularization every 16 steps with $\gamma=100$ but no path length regularization, and exponential moving average of model weights \cite{karras2017progressive}. We use the Adam optimizer \cite{adam} with learning rate $0.002$ and implement equalized learning rate for stability purposes as recommended by \cite{sgan2,karras2017progressive}.

We set $d_\text{noise} = 512$. Our layout generator $F$ is an 8-layer MLP with $d_\text{hidden} = 1024$ and leaky ReLU activations. We L2-normalize $\phi$ and $\psi$ vectors output by the layout generator before splatting. Altogether, the dimension of the last layer is $d_\text{out} = k(d_\text{in} + d_\text{style} + 5) + d_\text{in} + d_\text{style}$. To compensate for the removal of the first two convolutional blocks in the generator $G$, we increase channel widths at all remaining layers by 50\%. We set $d_\text{in}$ and $d_\text{style}$ depending on the number of blobs $k$, and values range between $256$ and $768$. We experiment with $k \in [5,50]$ depending on the data considered. We set the blob sigmoid temperature $c = 0.02$ by visual inspection of blob edge hardness. Model performance is relatively insensitive to jittering parameters. We perturb blob parameters with uniform noise as $\delta x \in [-0.04,0.04]$ (around $10$px at $256$px resolution), $\delta\theta \in [-0.1, 0.1] \ \text{rad}$ (around $6\degree$), and $\delta s \in [-0.5, 0.5]$ (varying radii of blobs by around 5px).

We train our model on categories from the LSUN scenes dataset \cite{yu2015lsun}. In particular, we train models on bedrooms; conference rooms; and the union of kitchens, living rooms, and dining rooms.  In the following section, we show results of models trained on bedroom data with $k=10$ blobs. Please see Appendix for results on more data \arxivoreccv{(\ref{app:otherdatasets})}{(\S A)}, further details on our blob parametrization and its implementation \arxivoreccv{(\ref{app:blobparam})}{(\S C)}, and ablations \arxivoreccv{(\ref{app:modelimpl})}{(\S E)}.

\begin{figure}[t]
\centering
\includegraphics[width=\textwidth]{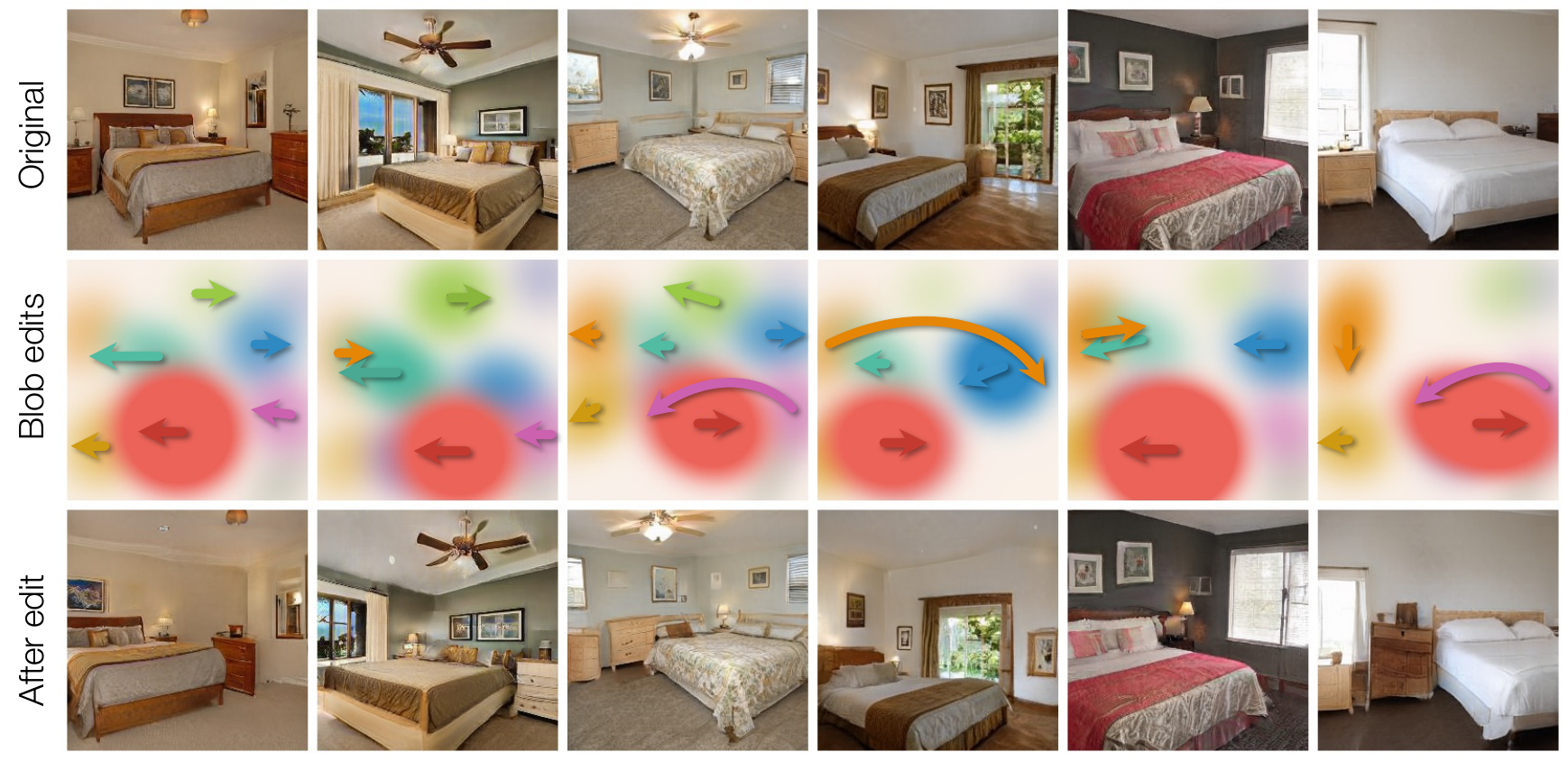}
\caption{\textbf{Moving blobs to rearrange objects:} By modifying coordinates $x_i$ of certain blobs as shown in the middle row, we can perform operations such as rearranging furniture in bedrooms. Note that since our representation is layered, we can model occlusions, such as the bed and the dresser in the leftmost and rightmost images.}
\label{fig:moveblobs}
\end{figure}

\begin{figure}[t]
\centering
\includegraphics[width=\textwidth]{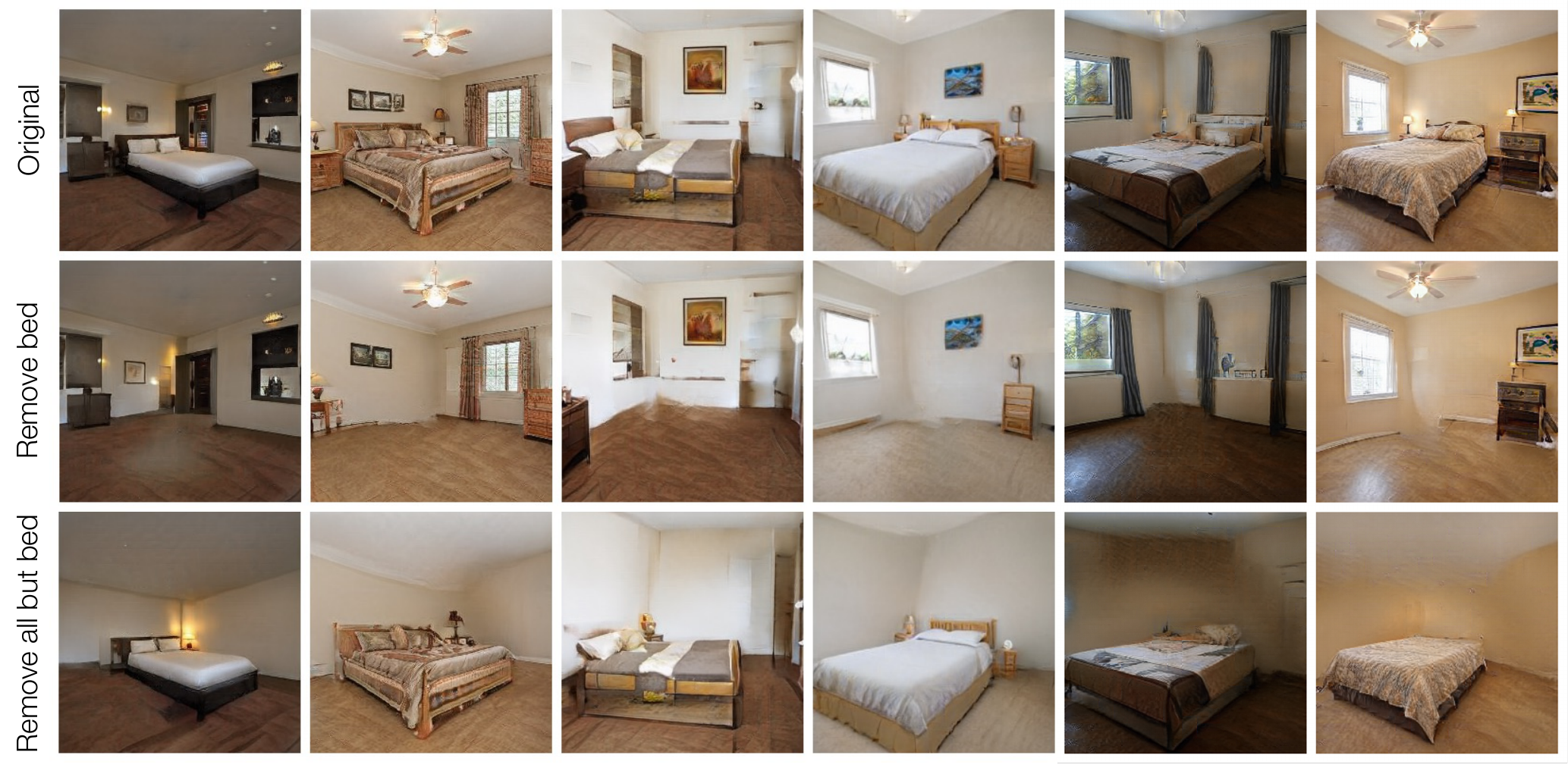}
\vspace{-4mm}
\caption{\textbf{Removing blobs:} Despite the extreme rarity of bedless bedrooms in the training data, the ability to remove beds from scenes by removing corresponding blobs emerges. We can also remove windows, lamps and fans, paintings, dressers, and nightstands in the same manner.}
\label{fig:removeblobs}
\vspace{-2mm}
\end{figure}

\begin{figure}[h]
\includegraphics[width=\textwidth]{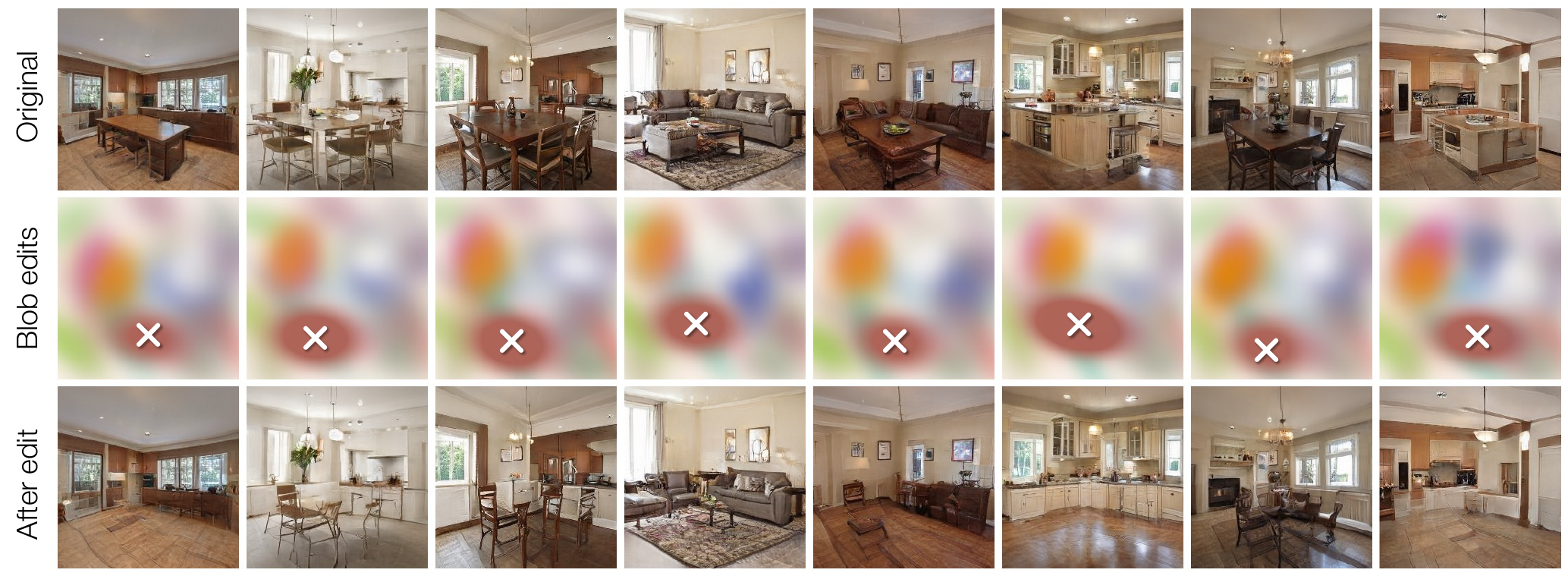}
\caption{\textbf{Removing all sorts of tables:} We train BlobGAN on a multi-category dataset of kitchens, living rooms, and dining rooms. We find that a particular blob specializes in generating tables across scene types, and feature vectors dictate whether it becomes a coffee table, kitchen island, or dining table. For many more editing operations on this dataset and others, please see Appendix.}
\label{fig:kld_remove_table}
\end{figure}

\subsection{Discovering entities with blobs}
The ideal representation is able to disentangle complex images of scenes into the objects that comprise them. Here, we demonstrate through various image manipulation applications that this ability emerges in our model. Our unsupervised representation allows effortless rearrangement, removal, cloning, and restyling of objects in scenes. We also measure correlation between blob presence and semantic categories as predicted by an off-the-shelf network and thus empirically verify the associations discovered by our model.

\begin{figure}[t]
\centering
\includegraphics[trim={0 10cm 0 0},clip,width=\textwidth]{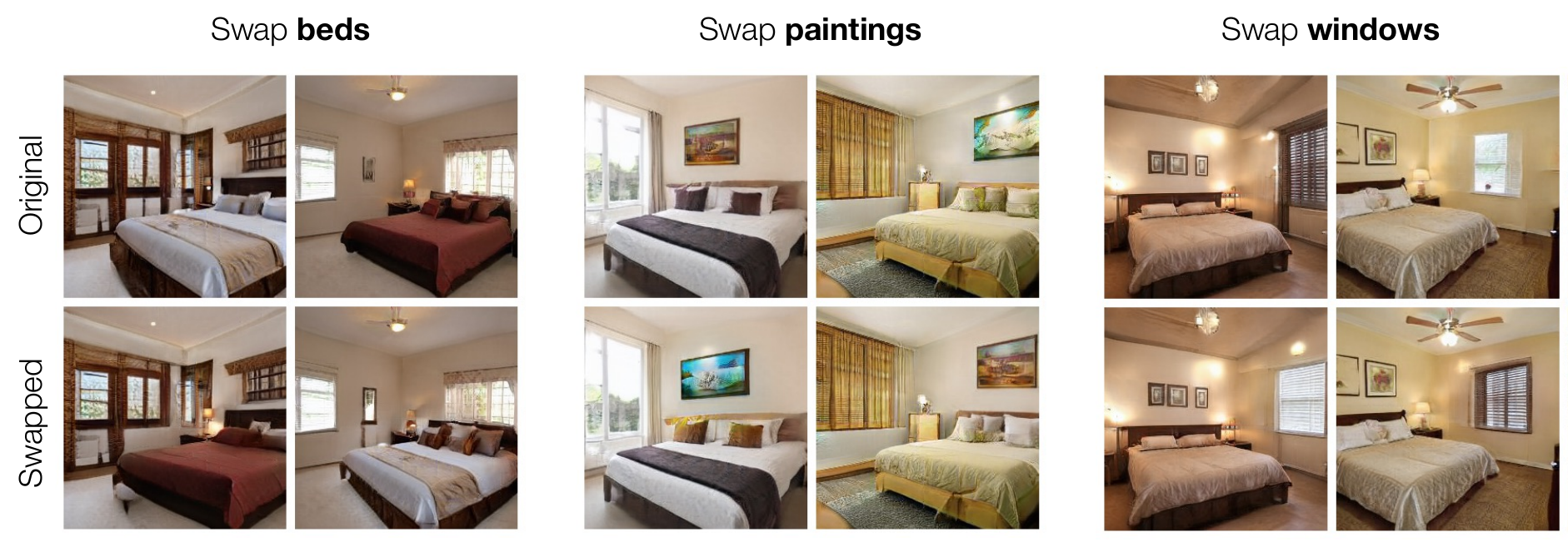}
\caption{\textbf{Swapping blob styles:} Interchanging $\psi_i$ vectors without modifying layout leads to localized edits which change the appearance of individual objects in the scene.}
\label{fig:localswap}
\end{figure}

Figure \ref{fig:moveblobs} shows the result of intervening to manipulate the center coordinates $x_i$ of blobs output by our model, and thus \textbf{rearranging furniture configurations}. We are able to arbitrarily alter the position of objects in the scene by shifting their corresponding blobs without affecting their appearance. This interaction is related to traditional ``image reshuffling'' where rearrangement of image content is done in pixel space~\cite{simakov2008bidir,barnes2009pm,rottshaham2019singan}. Our model's notion of depth ordering also allows us to easily de-occlude objects -- {\em e.g.} curtains, dressers, or nightstands -- that were hidden in the original images, while also enabling the introduction of new occlusions by moving one blob behind another.

In Figure \ref{fig:removeblobs}, we show the effect of \textbf{removing entirely certain blobs} from the representation. Specifically, we remove all blobs but the one responsible for beds, and show that our model is able to clear out the room accordingly. We also remove the bed blob but leave the rest of the room intact, showing a remarkable ability to create bedless bedrooms, despite training on a dataset of rooms with beds. Figure \ref{fig:sequenceedits} shows the effect of resizing blobs to change window size; see \arxivoreccv{\ref{app:blobsizeshape}}{Appendix A} for further results on changing blob size and shape. In Fig. \ref{fig:kld_remove_table}, we remove a blob that our model -- trained on a challenging multi-category union of scene datasets -- has learned to associate with tables across scene categories.

\begin{figure}[t]
\centering
\includegraphics[width=\textwidth]{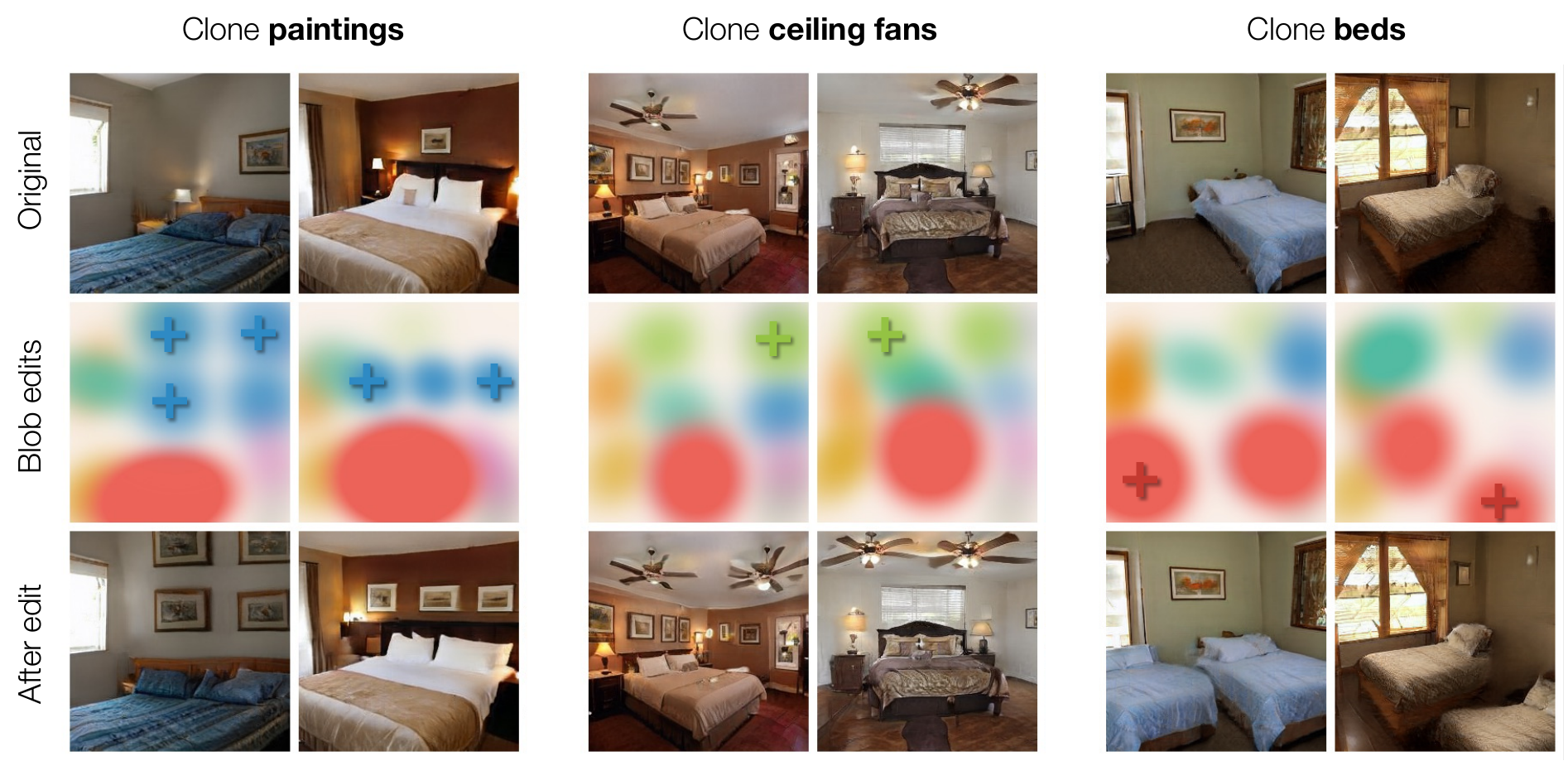}
\vspace{-.1in}
\caption{\textbf{Cloning blobs:} We clone blobs in scenes, arrange them to form a new layout, and show corresponding model outputs. Added blobs are marked with a $\boldsymbol{+}$.}
\label{fig:cloneblobs}
\end{figure}

Our edits are not constrained to the set of blobs present in a layout generated by our model; we can also introduce new blobs. Figure \ref{fig:cloneblobs} demonstrates the impact of \textbf{copying and pasting the same blob} in a new location. Our model is able to faithfully duplicate objects in scenes even when the duplication yields an image that is out of distribution, such as a room with two ceiling fans.

Our representation also allows performing edits across images. Figure \ref{fig:localswap} shows the \textbf{highly granular redecorating} enabled by swapping blob style vectors; we are able to copy objects such as bedsheets, windows, and artwork from one room to another without otherwise affecting the rendered scene.

\begin{figure}[t]
\centering
\includegraphics[width=\textwidth]{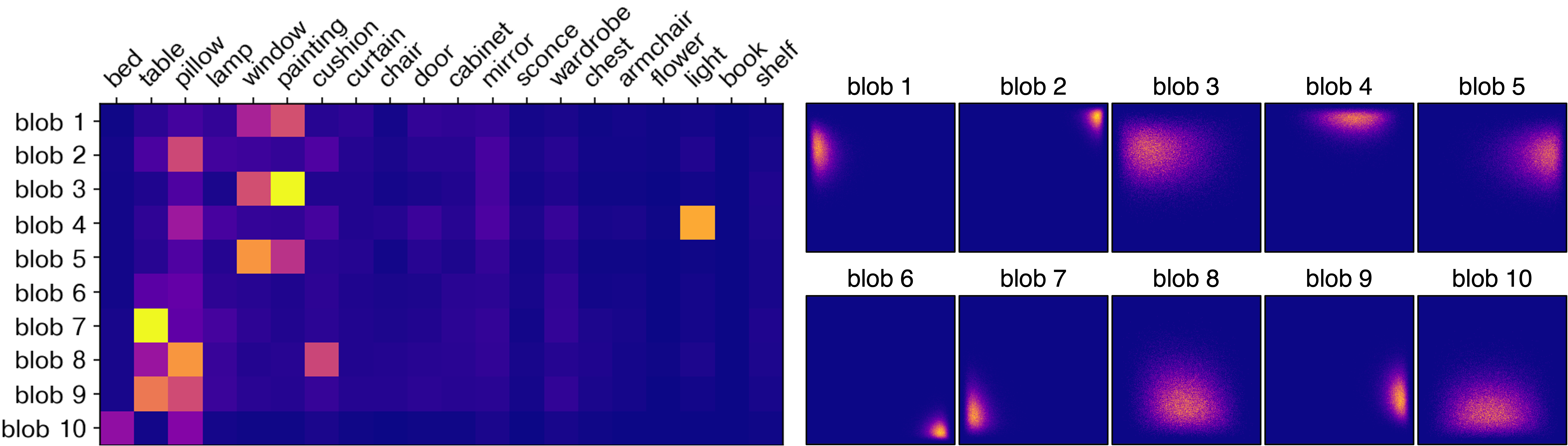}
\caption{\textbf{Blob spatial preferences:} Our model allocates each blob to a certain region of the image canvas, revealing patterns in the training distribution of objects. We visualize each blob's correlation with classes predicted by a segmentation model~\cite{strudel2021segmenter} {\em (left)} as well as the spatial distribution of blob centroids {\em (right)}. }
\label{fig:heatmap}
\end{figure}

\textbf{Quantitative blob analysis.} Next, we quantitatively study the strong associations between blobs and semantic object classes. We do so by randomly setting the size parameter $s$ of a blob to a large negative number to effectively remove it. We then use an off-the-shelf segmentation model to measure which semantic class has disappeared. We visualize the correlation between classes and blobs in Figure~\ref{fig:heatmap} \textit{(left)}; the sparsity of this matrix shows that blobs learn to specialize into distinct scene entities. We also visualize the distribution of blob centroids in Figure~\ref{fig:heatmap} \textit{(right)}, computed by sampling many different random vectors $z$. The resultant heatmaps provide a glimpse into the distribution of objects in training data -- our model learns to locate blobs at specific image regions and control the objects they represent by varying feature vectors.%

\subsection{Composing blobs into layouts}
\label{sec:layoutresults}
The ideal representation of scenes must go beyond simply disentangling images into their component parts, and capture the rich contextual relationships between these parts that dictate the process of scene formation \cite{biederman1981semantics,hock1978real}. In contrast to previous work in generative modeling of realistic images, our representation explicitly discovers the layout ({\em i.e.}, the joint distribution) of objects in scenes. 

By solving a simple constrained optimization problem at test-time, we are able to sample realistic images that satisfy constraints about the underlying scene, a functionality we call ``scene auto-complete''. This auto-complete allows applications such as filling empty rooms with items, plausibly populating rooms given a bed or window at a certain location, and finding layouts that are compatible with certain sets of furniture.

We ground this ability quantitatively by demonstrating that ``not everything goes with everything'' \cite{brooks2021hallucinating} in real-world scenes -- for example, not every room's style can be combined with any room's layout. We show that our scene auto-complete yields images that are significantly more photorealistic than na\"{i}vely combining scene properties at random, and outperforms regular StyleGAN in image quality, diversity as well as in fidelity of edits. %

\begin{figure}[t]
\centering
\includegraphics[width=\textwidth]{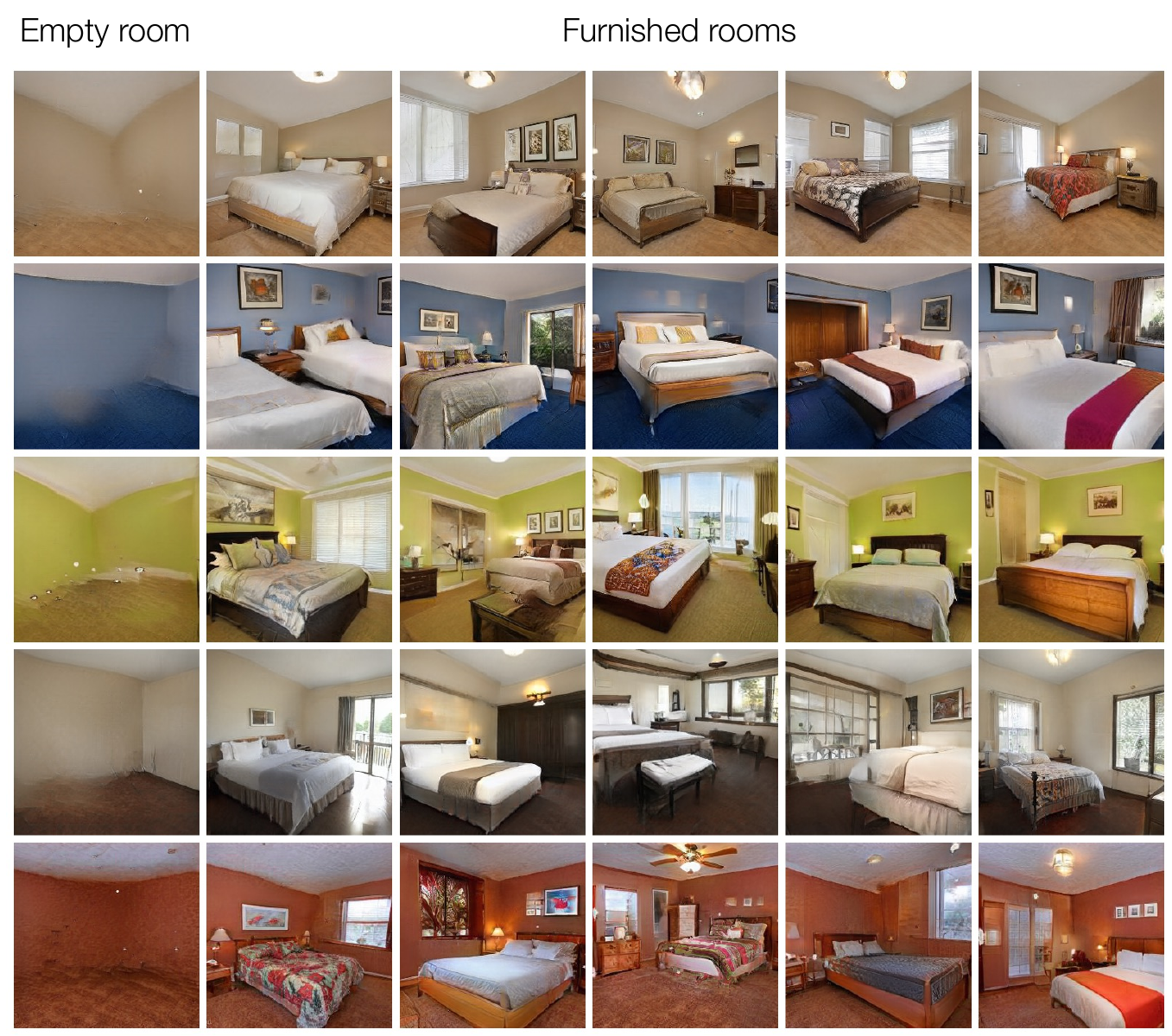}
\vspace{-.15in}
\caption{\textbf{Generating and populating empty rooms:}  We show different empty rooms, each with their own background vector $\psi_0$, as well as furnished rooms given by latents $z$ optimized to match these background vectors. This simple sampling procedure yields a diverse range of layouts to fill the scenes. Note that while empty rooms do not appear in training data, our model is reasonably capable of generating them.}
\vspace{-.05in}
\label{fig:emptyrooms}
\end{figure}

\begin{figure}[t!]
\centering
\includegraphics[trim={0 0 0 0},clip,width=\textwidth]{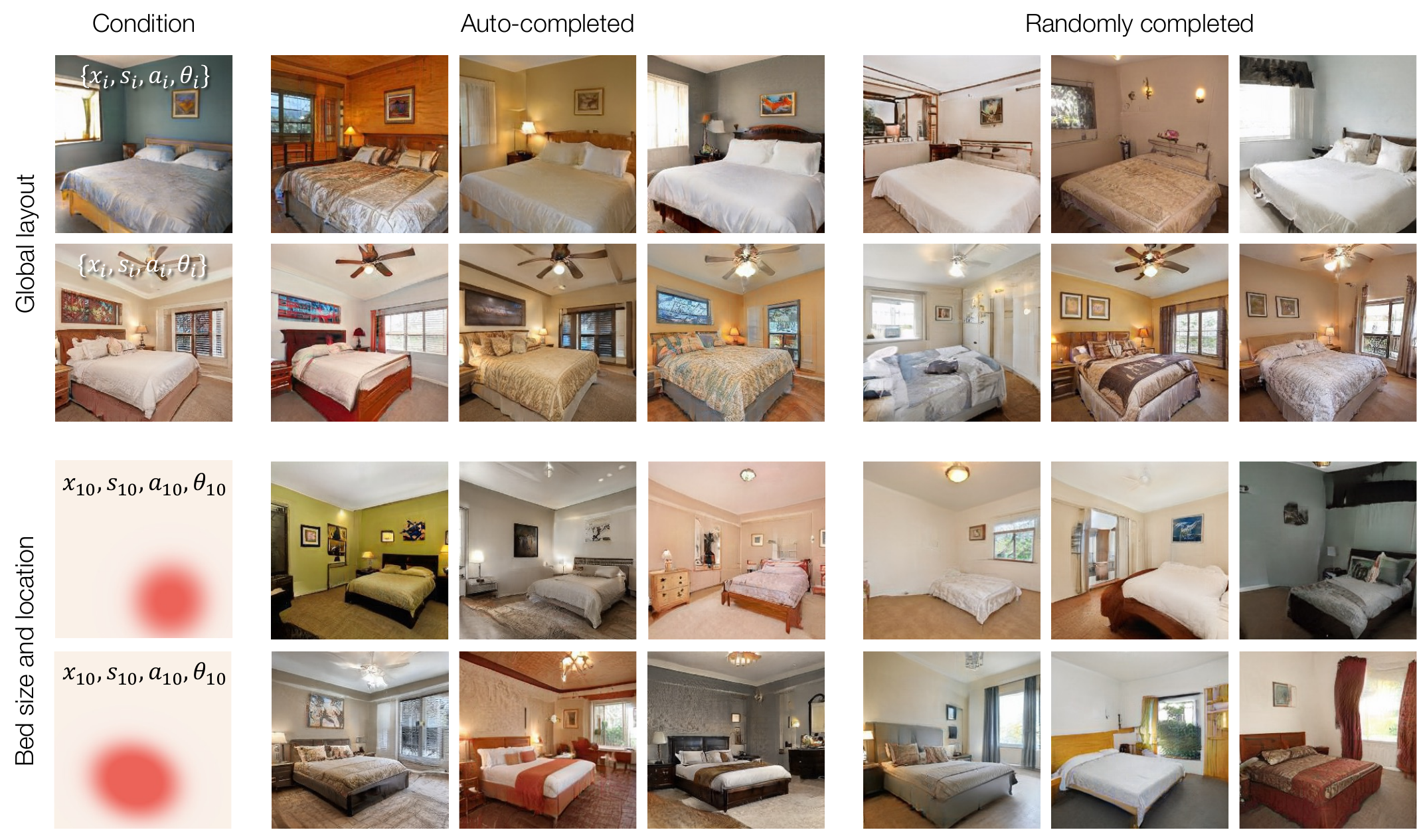}
\vspace{-.1in}
\caption{\textbf{Scene auto-complete:} Various conditional generation problems fall under the umbrella of ``scene auto-complete'', {\em i.e.} using our layout network $F$ to sample different scenes satisfying constraints on a subset of blob parameters.
We show layout-conditioned style generation as well as prediction of plausible scenes given the location and size (but not style) of beds. %
Rather than using $F$ to plausibly auto-complete scenes, we can also generate a random scene and simply override parameters of interest to match desired values. As shown on the right, such scenes have objects inserted, removed, reoriented, or otherwise disfigured due to incompatibility.}
\vspace{-.15in}
\label{fig:autocomplete}
\end{figure}
\textbf{Conditionally sampling scenes:} We can construct an ad-hoc conditional distribution by optimizing random inputs to match a set of constraints in the form of properties $c$ of a source image's blob map $\boldsymbol\beta$:
\begin{equation}
    c \subset \bigcup_{i=0}^{k} \{x_i^\text{src}, s_i^\text{src}, a_i^\text{src}, \theta_i^\text{src}, \phi_i^\text{src}, \psi_i^\text{src} \}
\end{equation}
For example, $c = \{\phi^\text{src}_0, \psi^\text{src}_0 \}$ constrains the background of an output image to match that of a source image, and $c = \{x^\text{src}_i, s^\text{src}_i, a^\text{src}_i\}$ constrains the shape (but not the appearance) of the $i$-th blob to match the source. 

We obtain conditional samples by drawing initial noise vectors $z^\text{init} \sim \mathcal{N}(0, I_d)$ and optimizing $F(z^\text{init})$ to match the constraint set $c$ with an L2 loss, leaving other parameters free. We use the Adam optimizer with learning rate $0.01$ and find that between 50 and 300 iterations, which complete in around a second on an NVIDIA RTX 3090, give $z^\text{optim}$ vectors that sufficiently match constraints. We then set the final layout to be $c \ \cup \ \{{\boldsymbol\beta}^\text{optim} \setminus c\}$, {\em i.e.} the initial constraints combined with the free parameters given by the optimized noise vectors, and decode layouts into images as described in Section \ref{sec:blobtoimg}. In effect, this process finds new scenes known by our model to be compatible with the specified constraints, as opposed to randomly drawn from an unconditional distribution.

\begin{table}[t]
  \centering
  \begin{adjustbox}{max width=\textwidth}
\begin{tabular}{ll|cccc|cccc|cccc|cccc}
\Xhline{2\arrayrulewidth}
\multicolumn{2}{c|}{} & \multicolumn{4}{c|}{Layout $\rightarrow$ Styles} & \multicolumn{4}{c|}{Window $\rightarrow$ Room} & \multicolumn{4}{c|}{Bed $\rightarrow$ Room} & \multicolumn{4}{c}{Painting $\rightarrow$ Room} \\ \hline
 \multicolumn{2}{r|}{} & \multicolumn{1}{c}{\makecell{FID  $\downarrow$}} & \multicolumn{1}{c}{\makecell{PD $\uparrow$}} & \multicolumn{1}{c}{\makecell{GD $\uparrow$}} & \multicolumn{1}{c|}{\makecell{C $\uparrow$}} & \multicolumn{1}{c}{\makecell{FID }} & \multicolumn{1}{c}{\makecell{PD }} & \multicolumn{1}{c}{\makecell{GD }} & \multicolumn{1}{c|}{\makecell{C }} & \multicolumn{1}{c}{\makecell{FID }} & \multicolumn{1}{c}{\makecell{PD  }} & \multicolumn{1}{c}{\makecell{GD }} & \multicolumn{1}{c|}{\makecell{C}} & \multicolumn{1}{c}{\makecell{FID }} & \multicolumn{1}{c}{\makecell{PD}} & \multicolumn{1}{c}{\makecell{GD }} & \multicolumn{1}{c}{\makecell{C}} \\ \hline

\multirow{3}{*}{\rotatebox[origin=c]{90}{\tiny StyleGAN}}   & 2 coarse & 4.23 & 0.75 & 0.77 & 46.5 &  - & - & - & - & - & - & - & - & - & - & - & -\\ 
 & 3 coarse & 5.04  & 0.73 & 0.76 & 55.3 & - & -& - & -  & - & - & - & - & - & - & - & -\\ 
 & 4 coarse & 5.58 & 0.71 & 0.76 & 62.9 &  - & - & - & - & - & - & - & - & - & - & - & -\\ 
 \hline
\multirow{3}{*}{\rotatebox[origin=c]{90}{\tiny BlobGAN}} & Random & 8.10 & 0.72 & 0.74 & 47.9 & 6.41 & 0.72 & 0.73 & 17.4 & 10.88 & 0.67 & 0.73 & 52.6 & 6.31 & 0.72 & 0.73 & 7.7 \\
 & Conditional & 4.59  & 0.70 & 0.74  & 55.2 &  {4.75} & {0.67} & {0.72} & {27.2}  & {7.12} & {0.64} & {0.72} & {60.0} & {4.58} & {0.69} & {0.73} & {13.0}   \\
 & {+ source $\Phi$} & {5.06}  & {0.68} & {0.74} & {63.6} & - &- & - & - & - & - & - & - & - & - & - & - \\
\Xhline{2\arrayrulewidth}
\end{tabular}
\end{adjustbox}
\vspace{0.5em}
\caption{\textbf{Not everything goes with everything:} We edit images by overriding properties in target images either generated at random or  conditionally sampled using our model. By varying the network depth at which we begin to swap styles in StyleGAN, we tune a knob between image quality and edit consistency. To further preserve global layout and improve consistency, our model can also use structure grids $\Phi$ from the source image. \textbf{PD} = paired distance, \textbf{GD} = global diversity, \textbf{C} = consistency. In all cases, scene auto-complete outperforms baselines.
Metrics are defined in the main text.}
\vspace{-.3in}
\label{tab:cond_fid}
\end{table}

We examine applications of our scene auto-complete and compare it to scenes generated by baseline approaches in Figures \ref{fig:emptyrooms} and \ref{fig:autocomplete}. Scene auto-complete yields images that are both more realistic and more faithful to the desired image operations. We quantitatively demonstrate this in Table \ref{tab:cond_fid}, where we show that using auto-complete to find target images whose properties to apply for conducting edits significantly outperforms the use of randomly sampled targets and/or models such as StyleGAN not trained with compositionality in mind. 

To evaluate image photorealism after an edit, we calculate FID \cite{heusel2017gans} on automatically edited images. We must also ensure that image quality does not come at the expense of sample diversity; to this end, we measure the average LPIPS \cite{zhang2018unreasonable} distance between images before and after the edit and refer to this as Paired Distance (\textbf{PD}). We also measure the expected distance between pairs of edited images to gauge whether edits cause perceptual mode collapse, and call this Global Diversity (\textbf{GD}). Finally, we confirm that our editing operations stay faithful to the conditioning provided. For predicting style from layout, we simply report the fraction of image pixels whose predicted class label as output by a segmentation network \cite{strudel2021segmenter} remains the same. For localized object edits, we report the intersection-over-union of the set of pixels whose prediction was the target class 
before and after edit. 
We refer to this metric as Consistency (\textbf{C}).

Our results verify the intuition that, {\em e.g.}, not every configuration of furniture can fit a bed at a given location. Please see \arxivoreccv{\ref{app:styleswap}}{Appendix D} for more results.

\arxivoreccv{\vspace{-4mm}}{\vspace{-4mm}}
\subsection{Evaluating visual quality and diversity}

\begin{wraptable}{br}{\arxivoreccv{0.5}{0.5}\textwidth}
\vspace{-8mm}
\begin{tabular}{l@{\hskip 2mm}c@{\hskip 2mm}c@{\hskip 2mm}c}
\toprule
     & FID $\downarrow$  & Precision $\uparrow$ & Recall $\uparrow$ \\ \midrule
StyleGAN2 & 3.85 & 0.5932 & 0.4492 \\
BlobGAN &3.43 & 0.5974 & 0.4463 \\ \bottomrule
\end{tabular}
\vspace{2mm}
\caption{BlobGAN achieves visual quality competitive with StyleGAN2~\cite{karras2019analyzing} on LSUN Bedrooms. Our samples are more realistic but capture less of the data distribution \cite{kynkaanniemi2019improved}, perhaps by rejecting unconventional or malformed scenes in the training data.}
\arxivoreccv{\vspace{-4mm}}{\vspace{-4mm}}
\label{tab:fid}
\end{wraptable}

Our model achieves perceptual realism competitive with previous work. In Table~\ref{tab:fid}, we report FID~\cite{heusel2017gans}
as well as improved precision and recall ~\cite{kynkaanniemi2019improved}, which capture realism and diversity of samples. Bedroom images generated by our model appear more realistic than StyleGAN's~\cite{karras2019style}, but less diverse. We hypothesize this is due to the design of our representation, which rejects strange scene configurations that cannot be modeled by blobs. When trained on the challenging union of multiple LSUN indoor scene categories, {\em BlobGAN outperforms StyleGAN2}, indicating an ability to scale to harder data. See \arxivoreccv{\ref{app:otherdatasets}}{Appendix A} for details.

\subsection{Parsing images into regions}
\label{sec:inv_main}
\begin{figure}[t]
\centering
\includegraphics[width=\textwidth]{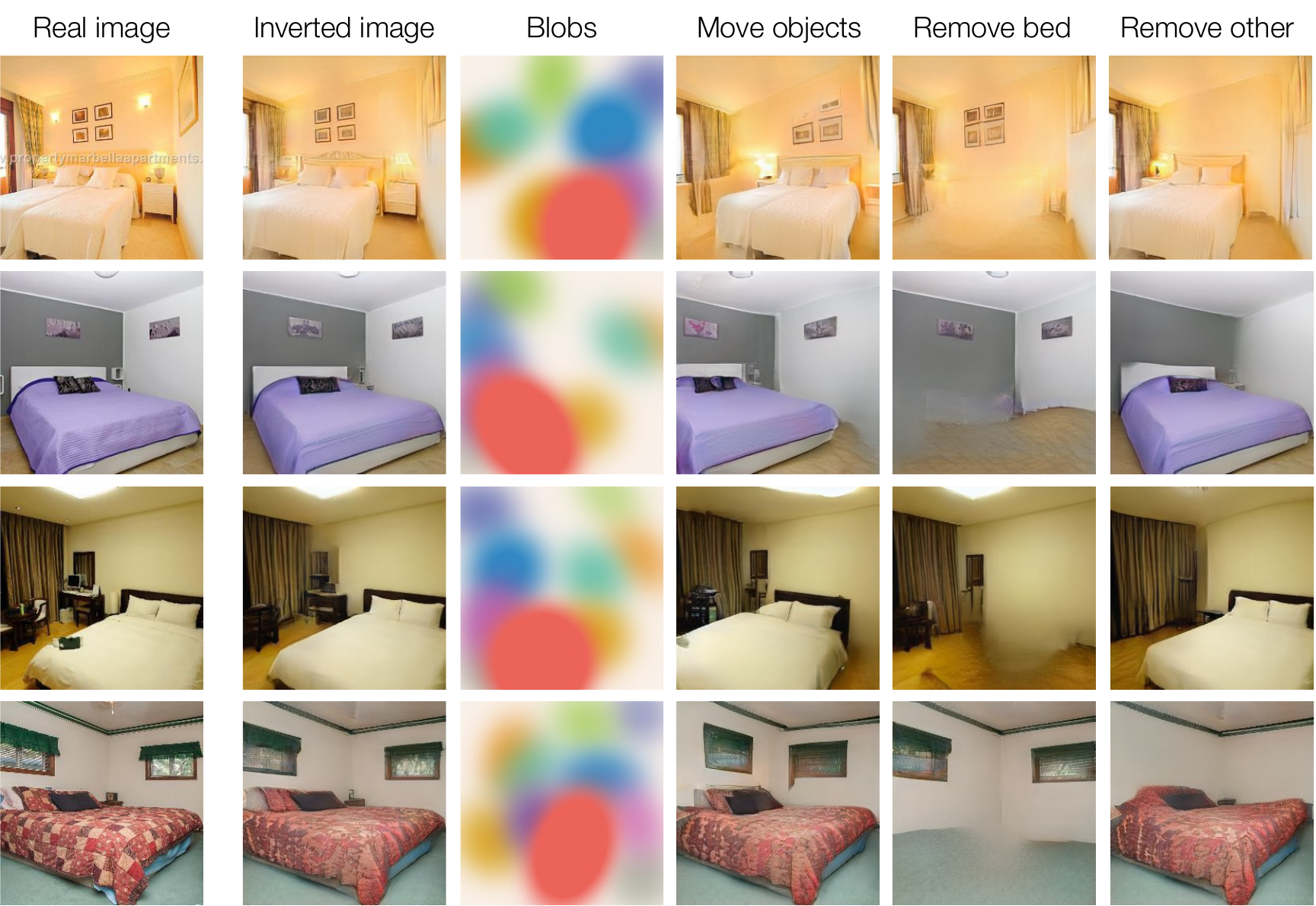}
\vspace{1mm}
\caption{\textbf{Parsing real images via inversion:}  Our representation can also parse real images by inverting them into blob space. We can remove and reposition objects in real images -- spot the differences from the original!}
\label{fig:inv}
\end{figure}

Though our representation is learned on generated ({\em i.e.} fake) images, in Figure \ref{fig:inv} we show that it can represent real images via inversion. We follow best practices~\cite{zhu2016generative,abdal2019image2stylegan,richardson2021encoding,tov2021designing,bau2019seeing} for inversion: We train an encoder
to predict blob parameters, reconstructing both real and fake images, and then
optimize encoder predictions to better reconstruct specific inputs. While this method leads to uneditable, off-manifold latents in previous work \cite{roich2021pivotal}, we find our blob representation to be more robust in this sense and amenable to na\"{i}ve optimization. Importantly, we find that the same manipulations described above can be readily applied to real images after inversion. See \arxivoreccv{\ref{app:inversion}}{Appendix B} for more information.

\section{Conclusion}

We present BlobGAN, a mid-level representation for generative modeling and parsing of scenes. Taking random noise as input, our model first outputs a set of spatial, depth-ordered blobs, and then splats these blobs onto a feature grid. This feature grid is used as input to a convolutional decoder which outputs images. While conceptually simple, this approach leads to the emergence of a disentangled representation that discovers entities in scenes and their layout. We demonstrate a set of edits enabled by our approach, including rearranging layouts by moving blobs and editing styles of individual objects. By removing or cloning blobs, we are even able to generate empty or densely populated rooms, though none exist in the training set. Our model can also parse and manipulate the layout of real images via inversion.

\arxivoreccv{}{}
\paragraph{Acknowledgements:}
We thank Allan Jabri, Assaf Shocher, Bill Peebles, Tim Brooks, and Yossi Gandelsman for endless insightful discussions and important feedback, and especially thank Vickie Ye for advice on blob compositing, splatting, and visualization. Thanks also to Georgios Pavlakos for deadline-week pixel inpsection and Shiry Ginosar for post-deadline-week guidance and helpful comments. Research was supported in part by the DARPA MCS program and a gift from Adobe Research. This work was started while DE was an intern at Adobe.

\bibliographystyle{splncs04}
\bibliography{egbib}

\begin{thebibliography}{100}
\providecommand{\url}[1]{\texttt{#1}}
\providecommand{\urlprefix}{URL }
\providecommand{\doi}[1]{https://doi.org/#1}

\bibitem{abdal2019image2stylegan}
Abdal, R., Qin, Y., Wonka, P.: Image2stylegan: How to embed images into the
  stylegan latent space? In: Proceedings of the IEEE/CVF International
  Conference on Computer Vision. pp. 4432--4441 (2019)

\bibitem{abdal2021styleflow}
Abdal, R., Zhu, P., Mitra, N.J., Wonka, P.: Styleflow: Attribute-conditioned
  exploration of stylegan-generated images using conditional continuous
  normalizing flows. ACM Transactions on Graphics (TOG)  \textbf{40}(3),  1--21
  (2021)

\bibitem{albahar2021pose}
AlBahar, B., Lu, J., Yang, J., Shu, Z., Shechtman, E., Huang, J.B.: Pose with
  {S}tyle: {D}etail-preserving pose-guided image synthesis with conditional
  stylegan. ACM Transactions on Graphics  (2021)

\bibitem{barnes2009pm}
Barnes, C., Shechtman, E., Finkelstein, A., Goldman, D.B.: Patchmatch: a
  randomized correspondence algorithm for structural image editing. {ACM}
  Trans. Graph.  \textbf{28}(3), ~24 (2009)

\bibitem{bau2020rewriting}
Bau, D., Liu, S., Wang, T., Zhu, J.Y., Torralba, A.: Rewriting a deep
  generative model. In: European Conference on Computer Vision. pp. 351--369.
  Springer (2020)

\bibitem{bau2018gan}
Bau, D., Zhu, J.Y., Strobelt, H., Zhou, B., Tenenbaum, J.B., Freeman, W.T.,
  Torralba, A.: Gan dissection: Visualizing and understanding generative
  adversarial networks. arXiv preprint arXiv:1811.10597  (2018)

\bibitem{bau2019seeing}
Bau, D., Zhu, J.Y., Wulff, J., Peebles, W., Strobelt, H., Zhou, B., Torralba,
  A.: Seeing what a gan cannot generate. In: Proceedings of the IEEE/CVF
  International Conference on Computer Vision. pp. 4502--4511 (2019)

\bibitem{bear2020learning}
Bear, D., Fan, C., Mrowca, D., Li, Y., Alter, S., Nayebi, A., Schwartz, J.,
  Fei-Fei, L.F., Wu, J., Tenenbaum, J., et~al.: Learning physical graph
  representations from visual scenes. Advances in Neural Information Processing
  Systems  \textbf{33},  6027--6039 (2020)

\bibitem{biederman1981semantics}
Biederman, I.: On the semantics of a glance at a scene (1981)

\bibitem{brock2018large}
Brock, A., Donahue, J., Simonyan, K.: Large scale gan training for high
  fidelity natural image synthesis (2018)

\bibitem{brooks2021hallucinating}
Brooks, T., Efros, A.A.: Hallucinating pose-compatible scenes. arXiv preprint
  arXiv:2112.06909  (2021)

\bibitem{carson1999blobworld}
Carson, C., Thomas, M., Belongie, S., Hellerstein, J.M., Malik, J.: Blobworld:
  A system for region-based image indexing and retrieval. In: International
  conference on advances in visual information systems. pp. 509--517. Springer
  (1999)

\bibitem{chai2021using}
Chai, L., Wulff, J., Isola, P.: Using latent space regression to analyze and
  leverage compositionality in gans. arXiv preprint arXiv:2103.10426  (2021)

\bibitem{chen2017photographic}
Chen, Q., Koltun, V.: Photographic image synthesis with cascaded refinement
  networks. In: Proceedings of the IEEE international conference on computer
  vision. pp. 1511--1520 (2017)

\bibitem{collins2020editing}
Collins, E., Bala, R., Price, B., Susstrunk, S.: Editing in style: Uncovering
  the local semantics of gans. In: Proceedings of the IEEE/CVF Conference on
  Computer Vision and Pattern Recognition. pp. 5771--5780 (2020)

\bibitem{denton2015deep}
Denton, E.L., Chintala, S., Fergus, R., et~al.: Deep generative image models
  using a￼ laplacian pyramid of adversarial networks. Advances in neural
  information processing systems  \textbf{28} (2015)

\bibitem{goetschalckx2019ganalyze}
Goetschalckx, L., Andonian, A., Oliva, A., Isola, P.: Ganalyze: Toward visual
  definitions of cognitive image properties. In: Proceedings of the ieee/cvf
  international conference on computer vision. pp. 5744--5753 (2019)

\bibitem{goodfellow2014generative}
Goodfellow, I., Pouget-Abadie, J., Mirza, M., Xu, B., Warde-Farley, D., Ozair,
  S., Courville, A., Bengio, Y.: Generative adversarial nets. Advances in
  neural information processing systems  \textbf{27} (2014)

\bibitem{gupta2010blocks}
Gupta, A., Efros, A.A., Hebert, M.: Blocks world revisited: Image understanding
  using qualitative geometry and mechanics. In: European Conference on Computer
  Vision. pp. 482--496. Springer (2010)

\bibitem{hariharan2014simultaneous}
Hariharan, B., Arbel{\'a}ez, P., Girshick, R., Malik, J.: Simultaneous
  detection and segmentation. In: European conference on computer vision. pp.
  297--312. Springer (2014)

\bibitem{harkonen2020ganspace}
H{\"a}rk{\"o}nen, E., Hertzmann, A., Lehtinen, J., Paris, S.: Ganspace:
  Discovering interpretable gan controls. Advances in Neural Information
  Processing Systems  \textbf{33},  9841--9850 (2020)

\bibitem{he2021masked}
He, K., Chen, X., Xie, S., Li, Y., Doll{\'a}r, P., Girshick, R.: Masked
  autoencoders are scalable vision learners. arXiv preprint arXiv:2111.06377
  (2021)

\bibitem{he2021latentkeypointgan}
He, X., Wandt, B., Rhodin, H.: Latentkeypointgan: Controlling gans via latent
  keypoints. arXiv preprint arXiv:2103.15812  (2021)

\bibitem{hedau2009recovering}
Hedau, V., Hoiem, D., Forsyth, D.: Recovering the spatial layout of cluttered
  rooms. In: 2009 IEEE 12th international conference on computer vision. pp.
  1849--1856. IEEE

\bibitem{heusel2017gans}
Heusel, M., Ramsauer, H., Unterthiner, T., Nessler, B., Hochreiter, S.: Gans
  trained by a two time-scale update rule converge to a local nash equilibrium.
  Advances in neural information processing systems  \textbf{30} (2017)

\bibitem{higgins2016beta}
Higgins, I., Matthey, L., Pal, A., Burgess, C., Glorot, X., Botvinick, M.,
  Mohamed, S., Lerchner, A.: beta-vae: Learning basic visual concepts with a
  constrained variational framework  (2016)

\bibitem{hock1978real}
Hock, H.S., Romanski, L., Galie, A., Williams, C.S.: Real-world schemata and
  scene recognition in adults and children. Memory \& Cognition  \textbf{6}(4),
   423--431 (1978)

\bibitem{hoiem2007ijcv}
Hoiem, D., Efros, A.A., Hebert, M.: Recovering surface layout from an image.
  IJCV  \textbf{75}(1),  151--172 (Oct 2007)

\bibitem{huang2018multimodal}
Huang, X., Liu, M.Y., Belongie, S., Kautz, J.: Multimodal unsupervised
  image-to-image translation. In: Proceedings of the European conference on
  computer vision (ECCV). pp. 172--189 (2018)

\bibitem{SceneCollaging}
Isola, P., Liu, C.: Scene collaging: Analysis and synthesis of natural images
  with semantic layers. In: ICCV (2013)

\bibitem{isola2017image}
Isola, P., Zhu, J.Y., Zhou, T., Efros, A.A.: Image-to-image translation with
  conditional adversarial networks. In: Proceedings of the IEEE conference on
  computer vision and pattern recognition. pp. 1125--1134 (2017)

\bibitem{jahanian2019steerability}
Jahanian, A., Chai, L., Isola, P.: On the" steerability" of generative
  adversarial networks. arXiv preprint arXiv:1907.07171  (2019)

\bibitem{johnson2018image}
Johnson, J., Gupta, A., Fei-Fei, L.: Image generation from scene graphs. In:
  Proceedings of the IEEE conference on computer vision and pattern
  recognition. pp. 1219--1228 (2018)

\bibitem{johnson2017clevr}
Johnson, J., Hariharan, B., Van Der~Maaten, L., Fei-Fei, L., Lawrence~Zitnick,
  C., Girshick, R.: Clevr: A diagnostic dataset for compositional language and
  elementary visual reasoning. In: Proceedings of the IEEE conference on
  computer vision and pattern recognition. pp. 2901--2910 (2017)

\bibitem{karras2017progressive}
Karras, T., Aila, T., Laine, S., Lehtinen, J.: Progressive growing of gans for
  improved quality, stability, and variation (2018)

\bibitem{karras2021alias}
Karras, T., Aittala, M., Laine, S., H{\"a}rk{\"o}nen, E., Hellsten, J.,
  Lehtinen, J., Aila, T.: Alias-free generative adversarial networks. Advances
  in Neural Information Processing Systems  \textbf{34} (2021)

\bibitem{karras2019style}
Karras, T., Laine, S., Aila, T.: A style-based generator architecture for
  generative adversarial networks. In: Proceedings of the IEEE/CVF conference
  on computer vision and pattern recognition. pp. 4401--4410 (2019)

\bibitem{sgan2}
Karras, T., Laine, S., Aittala, M., Hellsten, J., Lehtinen, J., Aila, T.:
  Analyzing and improving the image quality of stylegan. In: Proceedings of the
  IEEE/CVF conference on computer vision and pattern recognition. pp.
  8110--8119 (2020)

\bibitem{karras2019analyzing}
Karras, T., Laine, S., Aittala, M., Hellsten, J., Lehtinen, J., Aila, T.:
  Analyzing and improving the image quality of stylegan (2020)

\bibitem{adam}
Kingma, D.P., Ba, J.: Adam: A method for stochastic optimization. arXiv
  preprint arXiv:1412.6980  (2014)

\bibitem{kingma2013auto}
Kingma, D.P., Welling, M.: Auto-encoding variational bayes. arXiv preprint
  arXiv:1312.6114  (2013)

\bibitem{kynkaanniemi2019improved}
Kynk{\"a}{\"a}nniemi, T., Karras, T., Laine, S., Lehtinen, J., Aila, T.:
  Improved precision and recall metric for assessing generative models.
  Advances in Neural Information Processing Systems  \textbf{32} (2019)

\bibitem{lewis2021tryongan}
Lewis, K.M., Varadharajan, S., Kemelmacher-Shlizerman, I.: Tryongan: Body-aware
  try-on via layered interpolation. ACM Transactions on Graphics (Proceedings
  of ACM SIGGRAPH 2021)  \textbf{40}(4) (2021)

\bibitem{li2019diverse}
Li, K., Zhang, T., Malik, J.: Diverse image synthesis from semantic layouts via
  conditional imle. In: Proceedings of the IEEE/CVF International Conference on
  Computer Vision. pp. 4220--4229 (2019)

\bibitem{li2021collaging}
Li, Y., Li, Y., Lu, J., Shechtman, E., Lee, Y.J., Singh, K.K.: Collaging
  class-specific gans for semantic image synthesis. In: Proceedings of the
  IEEE/CVF International Conference on Computer Vision. pp. 14418--14427 (2021)

\bibitem{locatello2019challenging}
Locatello, F., Bauer, S., Lucic, M., Raetsch, G., Gelly, S., Sch{\"o}lkopf, B.,
  Bachem, O.: Challenging common assumptions in the unsupervised learning of
  disentangled representations. In: international conference on machine
  learning. pp. 4114--4124. PMLR (2019)

\bibitem{malisiewicz2009beyond}
Malisiewicz, T., Efros, A.: Beyond categories: The visual memex model for
  reasoning about object relationships. Advances in neural information
  processing systems  \textbf{22} (2009)

\bibitem{mejjati2021gaussigan}
Mejjati, Y.A., Milefchik, I., Gokaslan, A., Wang, O., Kim, K.I., Tompkin, J.:
  Gaussigan: Controllable image synthesis with 3d gaussians from unposed
  silhouettes. arXiv preprint arXiv:2106.13215  (2021)

\bibitem{mejjati2020objectstamps}
Mejjati, Y.A., Shen, Z., Snower, M., Gokaslan, A., Wang, O., Tompkin, J., Kim,
  K.I.: Generating object stamps. In: Computer Vision and Pattern Recognition
  Workshop on AI for Content Creation (CVPRW) (June 2020)

\bibitem{meng2021sdedit}
Meng, C., Song, Y., Song, J., Wu, J., Zhu, J.Y., Ermon, S.: Sdedit: Image
  synthesis and editing with stochastic differential equations. arXiv preprint
  arXiv:2108.01073  (2021)

\bibitem{nguyen2019hologan}
Nguyen-Phuoc, T., Li, C., Theis, L., Richardt, C., Yang, Y.L.: Hologan:
  Unsupervised learning of 3d representations from natural images. In:
  Proceedings of the IEEE/CVF International Conference on Computer Vision. pp.
  7588--7597 (2019)

\bibitem{nguyen2020blockgan}
Nguyen-Phuoc, T.H., Richardt, C., Mai, L., Yang, Y., Mitra, N.: Blockgan:
  Learning 3d object-aware scene representations from unlabelled images.
  Advances in Neural Information Processing Systems  \textbf{33},  6767--6778
  (2020)

\bibitem{glide}
Nichol, A., Dhariwal, P., Ramesh, A., Shyam, P., Mishkin, P., McGrew, B.,
  Sutskever, I., Chen, M.: Glide: Towards photorealistic image generation and
  editing with text-guided diffusion models. arXiv preprint arXiv:2112.10741
  (2021)

\bibitem{Niemeyer2020GIRAFFE}
Niemeyer, M., Geiger, A.: Giraffe: Representing scenes as compositional
  generative neural feature fields. In: Proc. IEEE Conf. on Computer Vision and
  Pattern Recognition (CVPR) (2021)

\bibitem{nitzberg19902}
Nitzberg, M., Mumford, D.B.: The 2.1-D sketch. IEEE Computer Society Press
  (1990)

\bibitem{Ohta-1978}
Ohta, Y., Kanade, T., Sakai, T.: An analysis system for scenes containing
  objects with substructures. In: Proceedings of 4th International Joint
  Conference on Pattern Recognition (IJCPR '78). pp. 752 -- 754 (November 1978)

\bibitem{oktay2018counterfactual}
Oktay, D., Vondrick, C., Torralba, A.: Counterfactual image networks (2018),
  \url{https://openreview.net/forum?id=SyYYPdg0-}

\bibitem{oliva2007role}
Oliva, A., Torralba, A.: The role of context in object recognition. Trends in
  cognitive sciences  \textbf{11}(12),  520--527 (2007)

\bibitem{park2019semantic}
Park, T., Liu, M.Y., Wang, T.C., Zhu, J.Y.: Semantic image synthesis with
  spatially-adaptive normalization. In: Proceedings of the IEEE/CVF conference
  on computer vision and pattern recognition. pp. 2337--2346 (2019)

\bibitem{park2020swapping}
Park, T., Zhu, J.Y., Wang, O., Lu, J., Shechtman, E., Efros, A., Zhang, R.:
  Swapping autoencoder for deep image manipulation. Advances in Neural
  Information Processing Systems  \textbf{33},  7198--7211 (2020)

\bibitem{patashnik2021styleclip}
Patashnik, O., Wu, Z., Shechtman, E., Cohen-Or, D., Lischinski, D.: Styleclip:
  Text-driven manipulation of stylegan imagery. In: Proceedings of the IEEE/CVF
  International Conference on Computer Vision. pp. 2085--2094 (2021)

\bibitem{pathakCVPR16context}
Pathak, D., Kr\"ahenb\"uhl, P., Donahue, J., Darrell, T., Efros, A.: Context
  encoders: Feature learning by inpainting (2016)

\bibitem{peebles2020hessian}
Peebles, W., Peebles, J., Zhu, J.Y., Efros, A., Torralba, A.: The hessian
  penalty: A weak prior for unsupervised disentanglement. In: European
  Conference on Computer Vision. pp. 581--597. Springer (2020)

\bibitem{alphacomp}
Porter, T., Duff, T.: Compositing digital images. In: Proceedings of the 11th
  annual conference on Computer graphics and interactive techniques. pp.
  253--259 (1984)

\bibitem{radford2015unsupervised}
Radford, A., Metz, L., Chintala, S.: Unsupervised representation learning with
  deep convolutional generative adversarial networks. arXiv preprint
  arXiv:1511.06434  (2015)

\bibitem{dalle2}
Ramesh, A., Dhariwal, P., Nichol, A., Chu, C., Chen, M.: Hierarchical
  text-conditional image generation with clip latents. arXiv preprint
  arXiv:2204.06125  (2022)

\bibitem{dalle}
Ramesh, A., Pavlov, M., Goh, G., Gray, S., Voss, C., Radford, A., Chen, M.,
  Sutskever, I.: Zero-shot text-to-image generation. In: International
  Conference on Machine Learning. pp. 8821--8831. PMLR (2021)

\bibitem{reed2016generative}
Reed, S., Akata, Z., Yan, X., Logeswaran, L., Schiele, B., Lee, H.: Generative
  adversarial text to image synthesis. In: International conference on machine
  learning. pp. 1060--1069. PMLR (2016)

\bibitem{richardson2021encoding}
Richardson, E., Alaluf, Y., Patashnik, O., Nitzan, Y., Azar, Y., Shapiro, S.,
  Cohen-Or, D.: Encoding in style: a stylegan encoder for image-to-image
  translation. In: Proceedings of the IEEE/CVF Conference on Computer Vision
  and Pattern Recognition. pp. 2287--2296 (2021)

\bibitem{roich2021pivotal}
Roich, D., Mokady, R., Bermano, A.H., Cohen-Or, D.: Pivotal tuning for
  latent-based editing of real images. arXiv preprint arXiv:2106.05744  (2021)

\bibitem{rombach2021high}
Rombach, R., Blattmann, A., Lorenz, D., Esser, P., Ommer, B.: High-resolution
  image synthesis with latent diffusion models. arXiv preprint arXiv:2112.10752
   (2021)

\bibitem{rottshaham2019singan}
Rott~Shaham, T., Dekel, T., Michaeli, T.: Singan: Learning a generative model
  from a single natural image. In: Computer Vision (ICCV), IEEE International
  Conference on (2019)

\bibitem{russell2009segmenting}
Russell, B., Efros, A., Sivic, J., Freeman, B., Zisserman, A.: Segmenting
  scenes by matching image composites. Advances in Neural Information
  Processing Systems  \textbf{22} (2009)

\bibitem{saharia2021palette}
Saharia, C., Chan, W., Chang, H., Lee, C.A., Ho, J., Salimans, T., Fleet, D.J.,
  Norouzi, M.: Palette: Image-to-image diffusion models. arXiv preprint
  arXiv:2111.05826  (2021)

\bibitem{sarkar2021style}
Sarkar, K., Golyanik, V., Liu, L., Theobalt, C.: Style and pose control for
  image synthesis of humans from a single monocular view (2021)

\bibitem{shelhamer2019blurring}
Shelhamer, E., Wang, D., Darrell, T.: Blurring the line between structure and
  learning to optimize and adapt receptive fields. arXiv preprint
  arXiv:1904.11487  (2019)

\bibitem{shen2020interfacegan}
Shen, Y., Yang, C., Tang, X., Zhou, B.: Interfacegan: Interpreting the
  disentangled face representation learned by gans. IEEE transactions on
  pattern analysis and machine intelligence  (2020)

\bibitem{shen2021closed}
Shen, Y., Zhou, B.: Closed-form factorization of latent semantics in gans. In:
  Proceedings of the IEEE/CVF Conference on Computer Vision and Pattern
  Recognition. pp. 1532--1540 (2021)

\bibitem{shi2000normalized}
Shi, J., Malik, J.: Normalized cuts and image segmentation. IEEE Transactions
  on pattern analysis and machine intelligence  \textbf{22}(8),  888--905
  (2000)

\bibitem{Siarohin_2019_NeurIPS}
Siarohin, A., Lathuilière, S., Tulyakov, S., Ricci, E., Sebe, N.: First order
  motion model for image animation. In: Conference on Neural Information
  Processing Systems (NeurIPS) (December 2019)

\bibitem{siarohin2021motion}
Siarohin, A., Woodford, O., Ren, J., Chai, M., Tulyakov, S.: Motion
  representations for articulated animation. In: CVPR (2021)

\bibitem{silberman2012indoor}
Silberman, N., Hoiem, D., Kohli, P., Fergus, R.: Indoor segmentation and
  support inference from rgbd images. In: European conference on computer
  vision. pp. 746--760. Springer (2012)

\bibitem{simakov2008bidir}
Simakov, D., Caspi, Y., Shechtman, E., Irani, M.: Summarizing visual data using
  bidirectional similarity. In: {CVPR}. {IEEE} Computer Society (2008)

\bibitem{strudel2021segmenter}
Strudel, R., Garcia, R., Laptev, I., Schmid, C.: Segmenter: Transformer for
  semantic segmentation. In: Proceedings of the IEEE/CVF International
  Conference on Computer Vision. pp. 7262--7272 (2021)

\bibitem{sudderth2005learning}
Sudderth, E.B., Torralba, A., Freeman, W.T., Willsky, A.S.: Learning
  hierarchical models of scenes, objects, and parts. In: Tenth IEEE
  International Conference on Computer Vision (ICCV'05) Volume 1. vol.~2, pp.
  1331--1338. IEEE (2005)

\bibitem{torralba2005describing}
Torralba, A., Willsky, A., Sudderth, E., Freeman, W.: Describing visual scenes
  using transformed dirichlet processes. Advances in neural information
  processing systems  \textbf{18} (2005)

\bibitem{tov2021designing}
Tov, O., Alaluf, Y., Nitzan, Y., Patashnik, O., Cohen-Or, D.: Designing an
  encoder for stylegan image manipulation. ACM Transactions on Graphics (TOG)
  \textbf{40}(4),  1--14 (2021)

\bibitem{tu2005image}
Tu, Z., Chen, X., Yuille, A.L., Zhu, S.C.: Image parsing: Unifying
  segmentation, detection, and recognition. International Journal of computer
  vision  \textbf{63}(2),  113--140 (2005)

\bibitem{wang2022improving}
Wang, J., Yang, C., Xu, Y., Shen, Y., Li, H., Zhou, B.: Improving gan
  equilibrium by raising spatial awareness. In: Proceedings of the IEEE/CVF
  Conference on Computer Vision and Pattern Recognition. pp. 11285--11293
  (2022)

\bibitem{wang2018high}
Wang, T.C., Liu, M.Y., Zhu, J.Y., Tao, A., Kautz, J., Catanzaro, B.:
  High-resolution image synthesis and semantic manipulation with conditional
  gans. In: Proceedings of the IEEE conference on computer vision and pattern
  recognition. pp. 8798--8807 (2018)

\bibitem{wu2021stylespace}
Wu, Z., Lischinski, D., Shechtman, E.: Stylespace analysis: Disentangled
  controls for stylegan image generation. In: Proceedings of the IEEE/CVF
  Conference on Computer Vision and Pattern Recognition. pp. 12863--12872
  (2021)

\bibitem{wulff2020improving}
Wulff, J., Torralba, A.: Improving inversion and generation diversity in
  stylegan using a gaussianized latent space. arXiv preprint arXiv:2009.06529
  (2020)

\bibitem{xia2021gan}
Xia, W., Zhang, Y., Yang, Y., Xue, J.H., Zhou, B., Yang, M.H.: Gan inversion: A
  survey. arXiv preprint arXiv:2101.05278  (2021)

\bibitem{YakimovskyF73}
Yakimovsky, Y., Feldman, J.A.: A semantics-based decision theory region
  analyser. In: {IJCAI}. pp. 580--588. William Kaufmann (1973)

\bibitem{yang2021semantic}
Yang, C., Shen, Y., Zhou, B.: Semantic hierarchy emerges in deep generative
  representations for scene synthesis. International Journal of Computer Vision
   \textbf{129}(5),  1451--1466 (2021)

\bibitem{yao20183d}
Yao, S., Hsu, T.M., Zhu, J.Y., Wu, J., Torralba, A., Freeman, B., Tenenbaum,
  J.: 3d-aware scene manipulation via inverse graphics. Advances in neural
  information processing systems  \textbf{31} (2018)

\bibitem{yu2015lsun}
Yu, F., Seff, A., Zhang, Y., Song, S., Funkhouser, T., Xiao, J.: Lsun:
  Construction of a large-scale image dataset using deep learning with humans
  in the loop. arXiv preprint arXiv:1506.03365  (2015)

\bibitem{yu2021unsupervised}
Yu, H.X., Guibas, L.J., Wu, J.: Unsupervised discovery of object radiance
  fields. arXiv preprint arXiv:2107.07905  (2021)

\bibitem{yu2002concurrent}
Yu, S.X., Gross, R., Shi, J.: Concurrent object recognition and segmentation by
  graph partitioning. Advances in neural information processing systems
  \textbf{15} (2002)

\bibitem{zhang2021decorating}
Zhang, C., Xu, Y., Shen, Y.: Decorating your own bedroom: Locally controlling
  image generation with generative adversarial networks. arXiv preprint
  arXiv:2105.08222  (2021)

\bibitem{zhang2019self}
Zhang, H., Goodfellow, I., Metaxas, D., Odena, A.: Self-attention generative
  adversarial networks. In: International conference on machine learning. pp.
  7354--7363. PMLR (2019)

\bibitem{zhang2017stackgan}
Zhang, H., Xu, T., Li, H., Zhang, S., Wang, X., Huang, X., Metaxas, D.N.:
  Stackgan: Text to photo-realistic image synthesis with stacked generative
  adversarial networks. In: Proceedings of the IEEE international conference on
  computer vision. pp. 5907--5915 (2017)

\bibitem{zhang2018unreasonable}
Zhang, R., Isola, P., Efros, A.A., Shechtman, E., Wang, O.: The unreasonable
  effectiveness of deep features as a perceptual metric. In: Proceedings of the
  IEEE conference on computer vision and pattern recognition. pp. 586--595
  (2018)

\bibitem{zhu2022region}
Zhu, J., Shen, Y., Xu, Y., Zhao, D., Chen, Q.: Region-based semantic
  factorization in gans. arXiv preprint arXiv:2202.09649  (2022)

\bibitem{zhu2020domain}
Zhu, J., Shen, Y., Zhao, D., Zhou, B.: In-domain gan inversion for real image
  editing. In: European conference on computer vision. pp. 592--608. Springer
  (2020)

\bibitem{zhu2016generative}
Zhu, J.Y., Kr{\"a}henb{\"u}hl, P., Shechtman, E., Efros, A.A.: Generative
  visual manipulation on the natural image manifold. In: European conference on
  computer vision. pp. 597--613. Springer (2016)

\bibitem{zhu2017toward}
Zhu, J.Y., Zhang, R., Pathak, D., Darrell, T., Efros, A.A., Wang, O.,
  Shechtman, E.: Toward multimodal image-to-image translation. Advances in
  neural information processing systems  \textbf{30} (2017)

\end{thebibliography}
\renewcommand{\theHchapter}{A\arabic{chapter}}
\clearpage
\renewcommand{\thesection}{Appendix~\Alph{section}}
\renewcommand{\thesubsection}{\Alph{section}.\arabic{subsection}}
\setcounter{section}{0}

\begin{figure}[h]
\centering
\includegraphics[width=\textwidth]{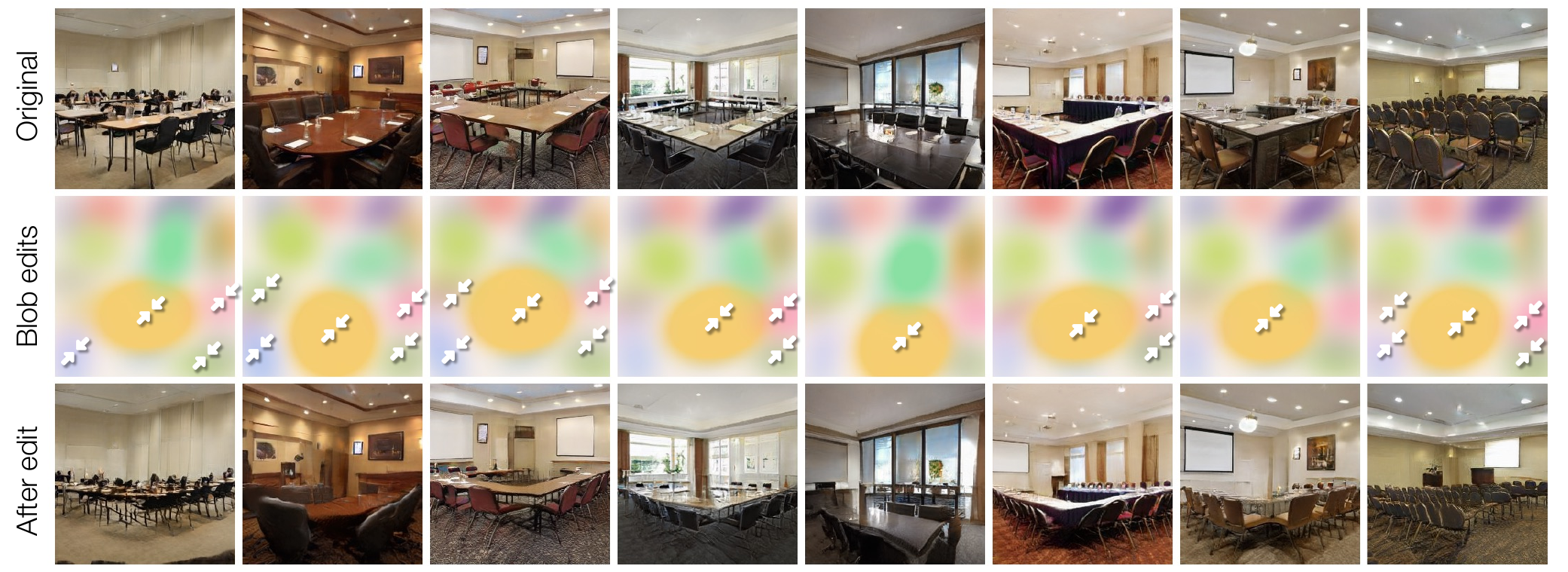}
\caption{\textbf{Honey, I shrunk the conference room!} As in Figure \ref{fig:sequenceedits}, we show the effect of resizing blobs in generated images. Here, we resize blobs corresponding to tables and chairs, and render identical rooms with shrunken furniture.}
\label{fig:conf_clear_floors}
\arxivoreccv{}{\vspace{-2em}}
\end{figure}

\begin{figure}[h]
\includegraphics[width=\textwidth]{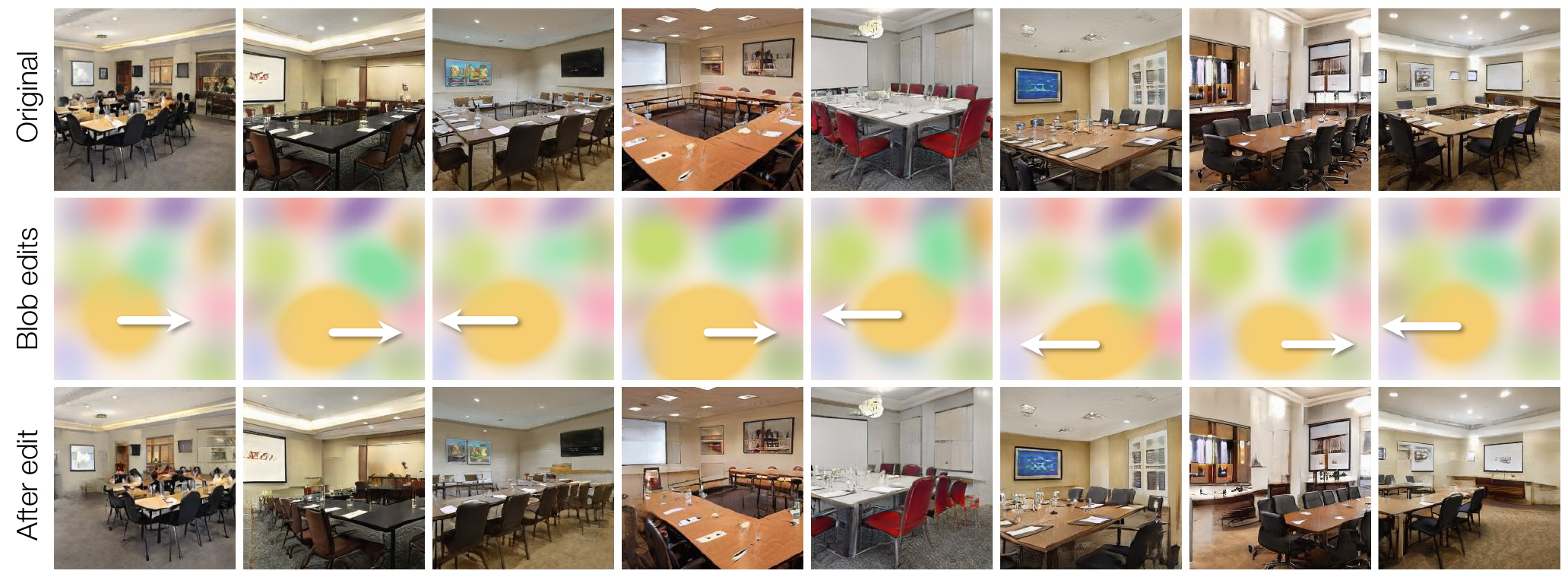}
\caption{\textbf{Moving desks and chairs (conference room):} As in Figure \ref{fig:moveblobs}, we show the effect of moving blobs in generated images. Here, we move blobs corresponding to tables and chairs, and render identical rooms with shifted furniture.}
\label{fig:conf_move_tables}
\end{figure}

\begin{figure}[h]
\arxivoreccv{}{\vspace{-2mm}}
\includegraphics[width=\textwidth]{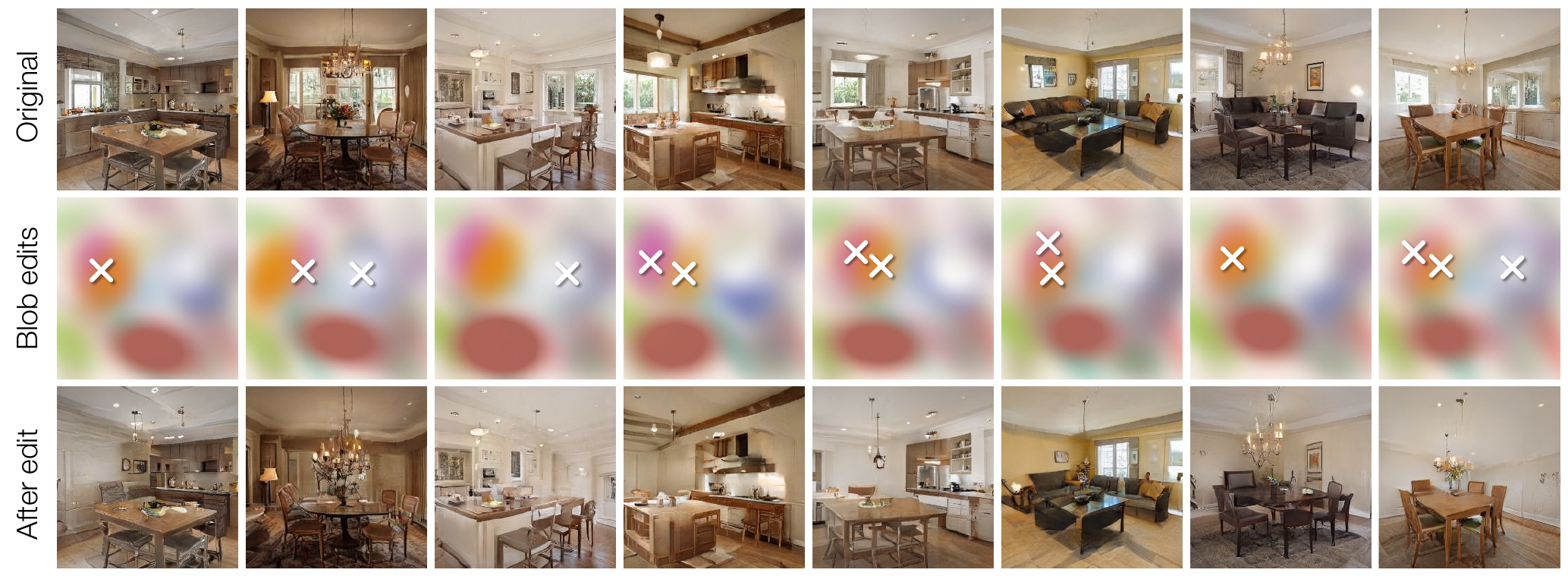}
\caption{\textbf{Removing some or all windows (kitchen, living room, dining room):} As shown in Figure \ref{fig:removeblobs}, we can remove windows from complex scenes, though they are often hidden behind cluttered configurations of furniture. We can control which windows to remove by selecting only some of the relevant blobs.}

\label{fig:kld_remove_window}
\end{figure}

\begin{figure}[h]
\includegraphics[width=\textwidth]{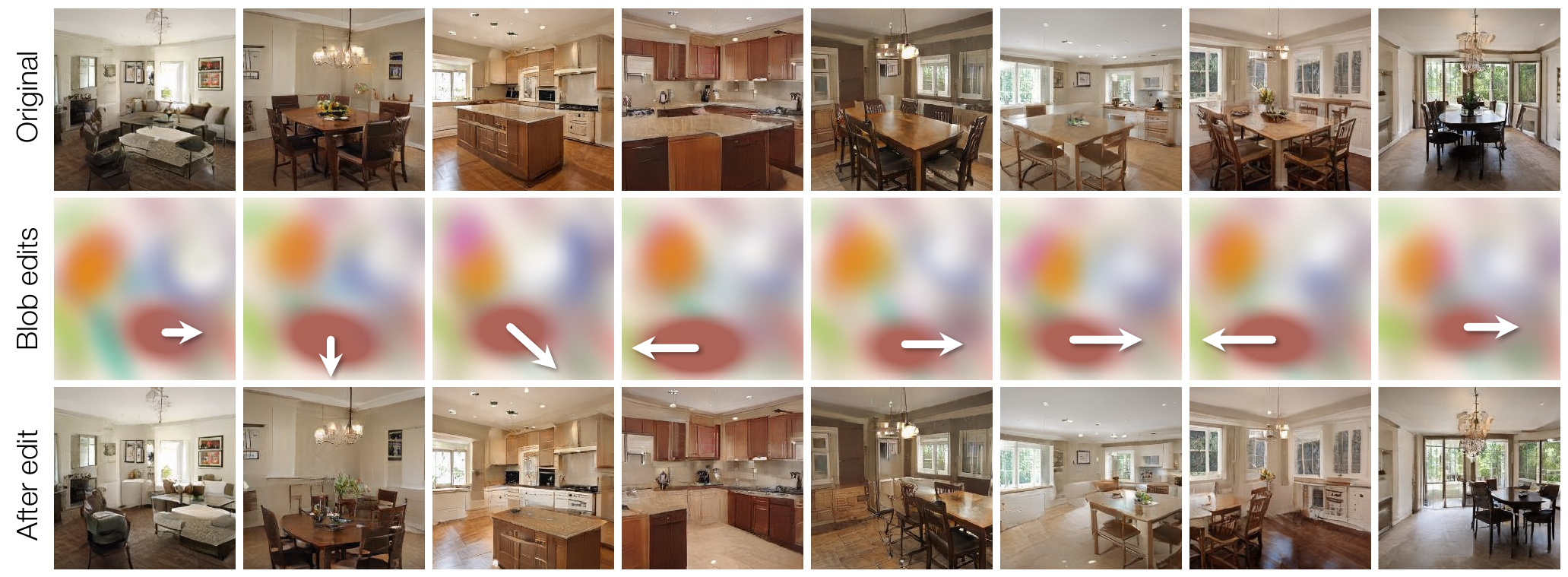}
\caption{\textbf{Moving tables and chairs (kitchen, living room, dining room):} Our representation can easily move tables and any associated chairs, by changing the location of blobs 42 (table) and 30 (chairs). Since the two move together, we only show one arrow to represent the edit.}

\label{fig:kld_move_table}
\end{figure}

\begin{figure}[h!]
\centering
\includegraphics[width=\textwidth]{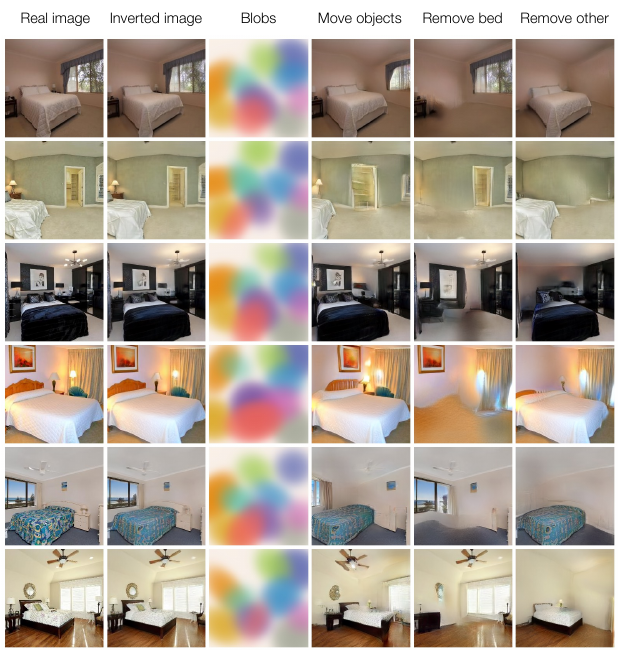}
\caption{\textbf{Parsing real images via inversion:}  We show the flexibility of our learned representation by applying edits to real images inverted into blob space. We can remove and reposition objects in real images -- spot the differences from the original!}
\label{fig:moreinversion}
\arxivoreccv{\vspace{-3mm}}{\vspace{-1em}}
\end{figure}

\begin{figure}[t]
\centering
\includegraphics[width=\textwidth]{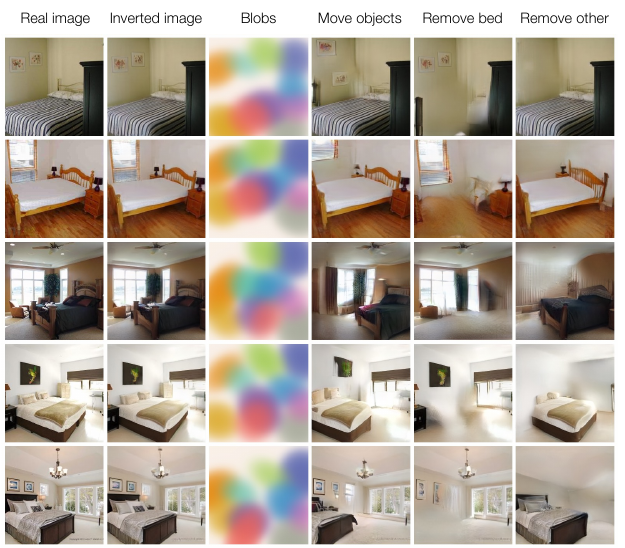}
\caption{\textbf{Parsing real images via inversion:} More results on inversion of real images.}
\label{fig:moreinversion2}
\arxivoreccv{\vspace{-3mm}}{}
\end{figure}

 \begin{figure}[h]
\includegraphics[width=\textwidth]{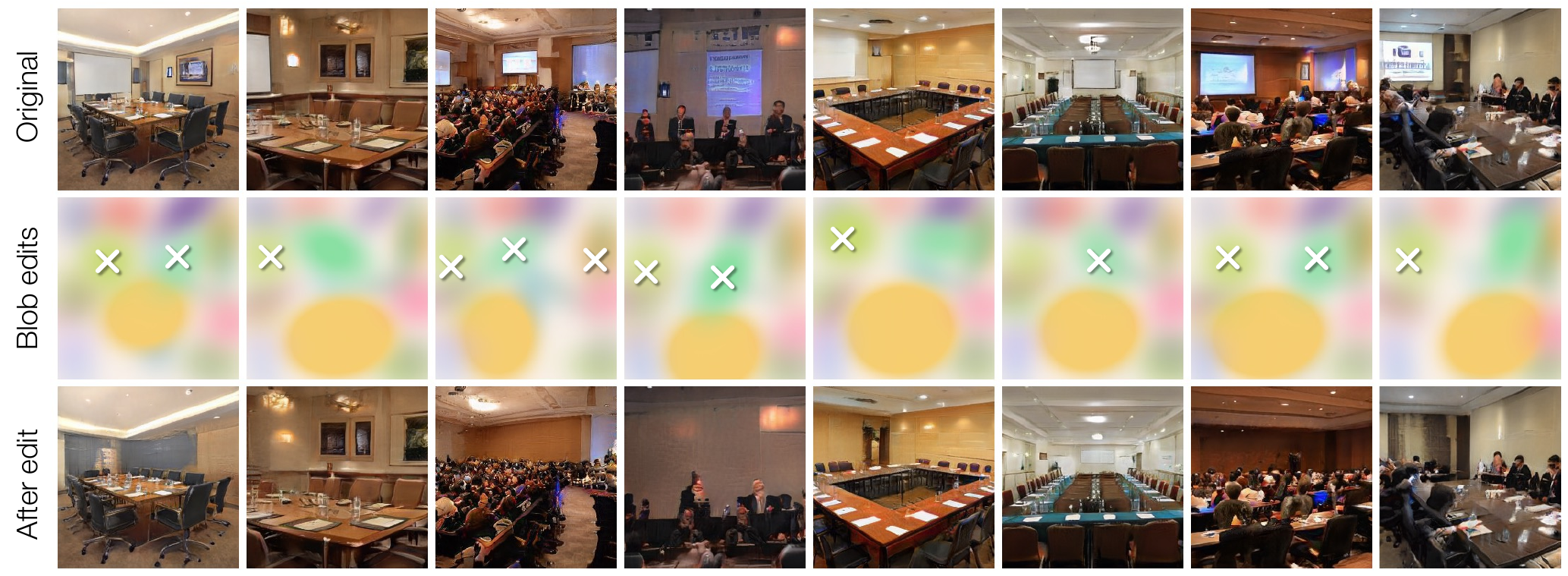}
\caption{\textbf{Removing screens in conference rooms.} As in Figure \ref{fig:removeblobs}, we show the effect of removing certain blobs from generated images. Here, we remove blobs corresponding to screens from images of conference rooms.}
\label{fig:conf_remove_screens}
\end{figure}

\begin{figure}[t]
\centering
\includegraphics[width=\textwidth]{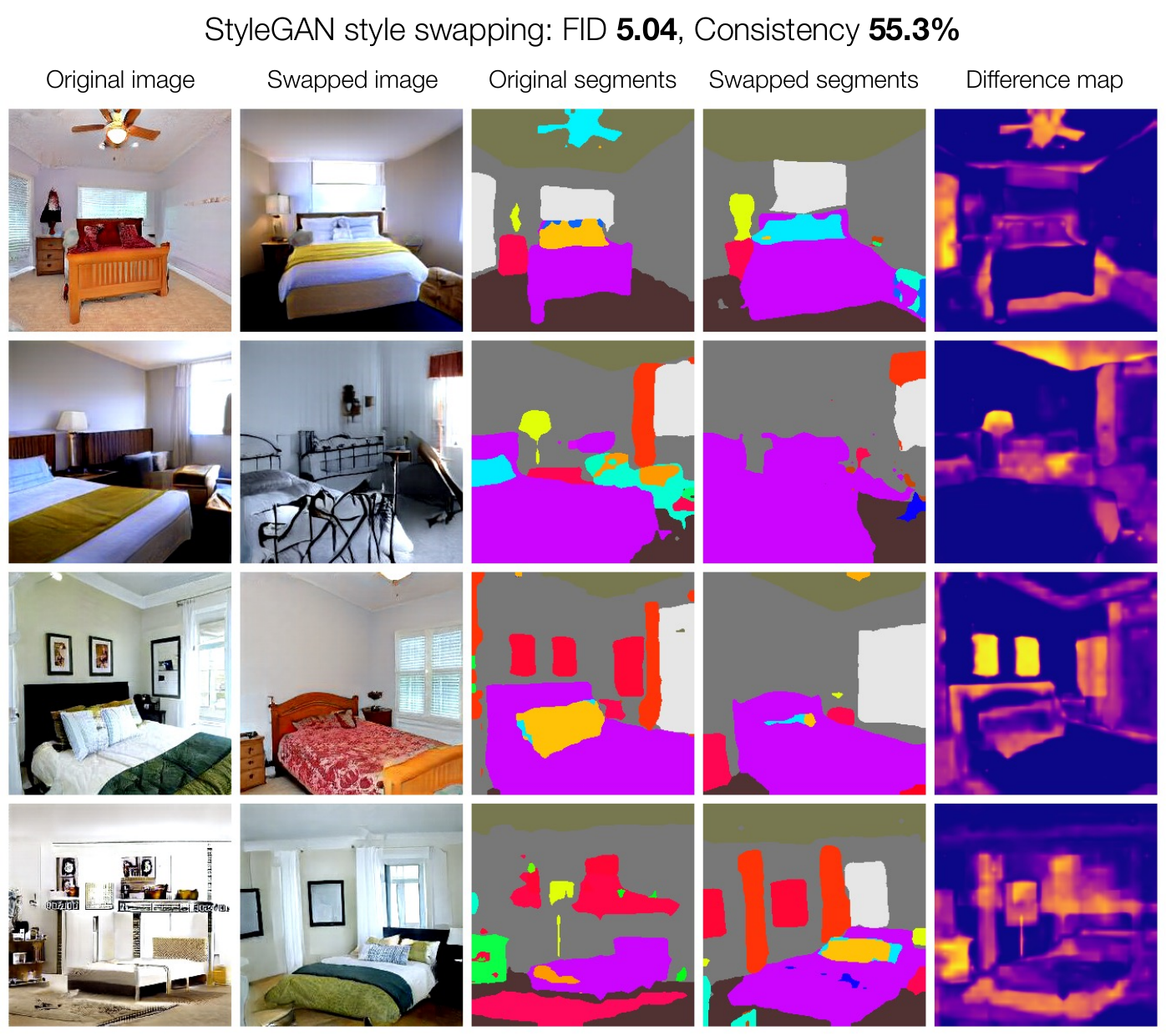}
\caption{\textbf{Style swapping with StyleGAN:} We show randomly sampled untruncated StyleGAN images before and after style swapping at layer 4, attaining an FID of 5.04 and layout consistency of 55.3\%. 
The difference map shows the normalized KL divergence of the predicted per-pixel logits before and after swapping.}
\label{fig:sganswap}
\end{figure}

\begin{figure}[t]
\centering
\includegraphics[width=\textwidth]{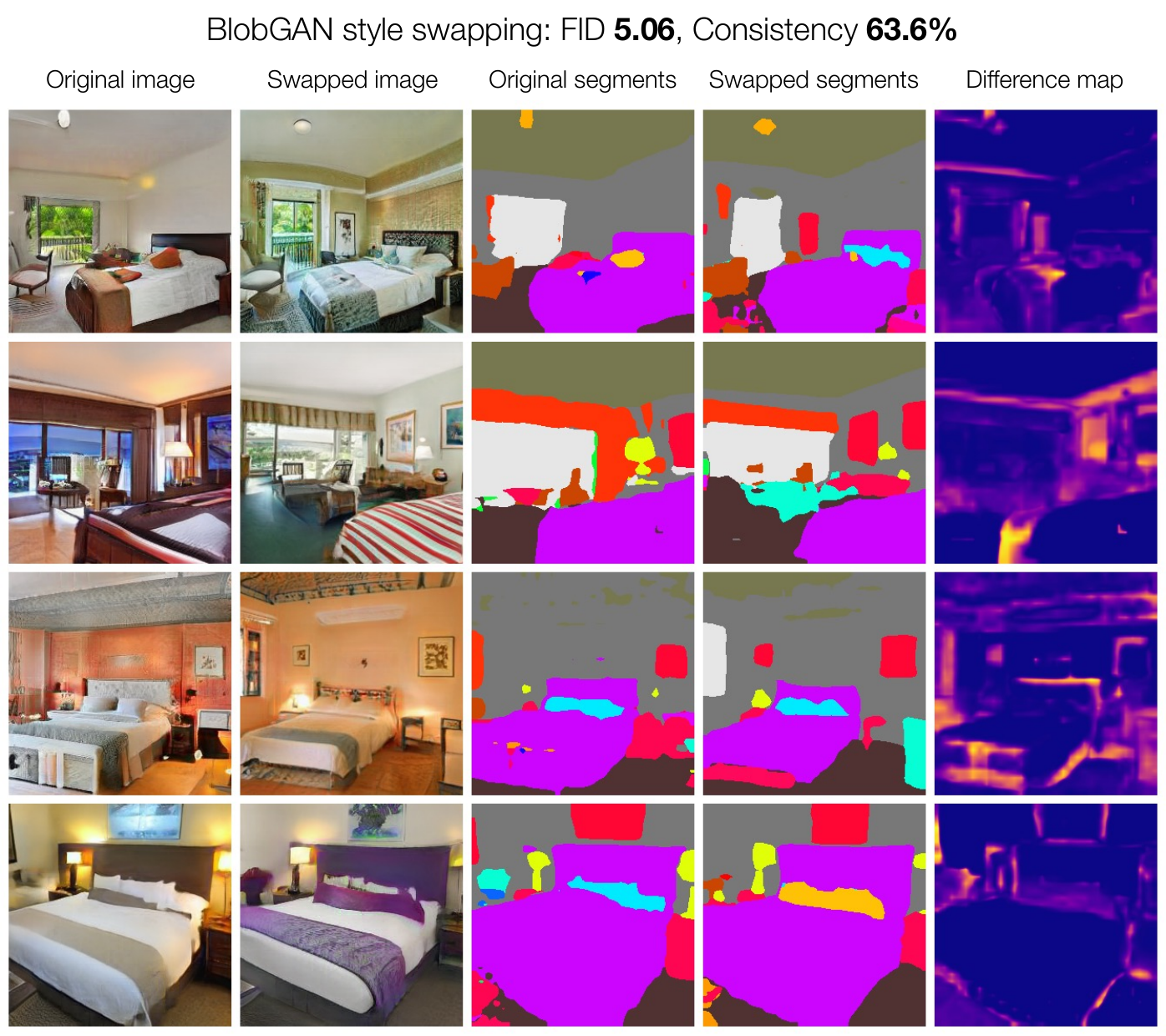}
\caption{\textbf{Style swapping with BlobGAN:} We show randomly sampled untruncated BlobGAN images before and after style swapping, attaining an FID of 5.06 and layout consistency of 63.6\%. 
The difference map shows the normalized KL divergence of the predicted per-pixel logits before and after swapping.}
\label{fig:blobswap}
\end{figure}

\section{BlobGAN on other datasets}
\label{app:otherdatasets}
\label{app:blobsizeshape}
\begin{figure}[h]
\centering

\includegraphics[trim={0 30.5cm 0 0},clip,width=\textwidth]{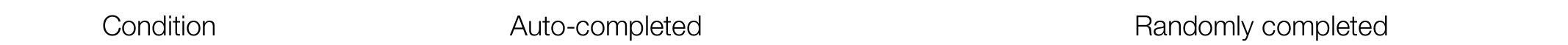}
\includegraphics[trim={0 0 0 22cm},clip,width=\textwidth]{figs/autocomplete_blobs.pdf}

\caption{\textbf{Scene auto-complete:} Various conditional generation problems fall under the umbrella of ``scene auto-complete'', {\em i.e.} optimizing random noise vectors to match a set of blob parameters when run through our layout network $F$. We show prediction of plausible scenes given the location and size (but not style) of dressers and nightstands. The model must not only predict the arrangement of the missing blobs, but also assign all blobs realistic appearance. When sampling target images randomly, objects are often randomly inserted, removed, reoriented, or otherwise disfigured due to incompatibility.}
\label{fig:autocomplete_dresser}
\end{figure}

In the main text, we primarily showed results on LSUN bedrooms~\cite{yu2015lsun}. Below, we show that our model can be applied to other datasets and room types. We provide qualitative and quantitative results on our models trained on the challenging LSUN conference room dataset, as well as a joint dataset combining LSUN kitchens, dining rooms, and living rooms \cite{yu2015lsun}. As with bedrooms, our model's images are competitive with previous work in terms of photorealism, and in addition allow extensive manipulation of images. Please see Table \ref{tab:fid_more} for quantitative evaluation. We show image samples and edits on them in Figures \ref{fig:conf_remove_screens}, \ref{fig:conf_clear_floors}, \ref{fig:conf_move_tables}, \ref{fig:kld_remove_window}, \ref{fig:localswap_row2}, \ref{fig:autocomplete_dresser}, and \ref{fig:kld_move_table}.
\begin{table}[h]
\vspace{5mm}
\centering
\begin{tabular}{l@{\hskip 2mm}c@{\hskip 2mm}c@{\hskip 2mm}c@{\hskip 2mm}c@{\hskip 2mm}c@{\hskip 2mm}c}
\toprule
& \multicolumn{3}{c}{LSUN Conference} & \multicolumn{3}{c}{LSUN Kitchen+Living+Dining} \\
    \cmidrule(lr){2-4} \cmidrule(lr){5-7}

     & FID $\downarrow$  & Precision $\uparrow$ & Recall $\uparrow$     & FID $\downarrow$  & Precision $\uparrow$ & Recall $\uparrow$ \\ \midrule
StyleGAN2 \cite{karras2019analyzing} & 6.21 & 0.5475 & 0.4554 & 4.63 & 0.6005 & 0.4397 \\
BlobGAN & 6.94 & 0.5297 & 0.4485 & 4.41 & 0.5818 & 0.4661\\

\bottomrule
\end{tabular}
\vspace{2mm}
\caption{On challenging collections of conference rooms and various types of indoor rooms in homes, our model is highly competitive with a StyleGAN2 baseline, while enabling all the applications of the BlobGAN representation. Our model outperforms StyleGAN2 given an equal number of gradient steps (1.5M) on the difficult union of various LSUN indoor scene categories, as measured by FID.}
\arxivoreccv{\vspace{-5mm}}{}
\label{tab:fid_more}
\end{table}

\section{Modeling real images with BlobGAN}

\label{app:inversion}
We show additional results on inversion and editing of real images in Figures \ref{fig:moreinversion} and \ref{fig:moreinversion2}. Images are drawn from the LSUN bedrooms validation set, which our model does not see during the training process.

\subsection{Implementation details} In Section \ref{sec:inv_main} and Figure \ref{fig:inv}, we demonstrate that real images can be inverted and manipulated with our model. Here, we provide additional details regarding the encoder training procedure. We take an encoder architecture $E$ in the same form as the StyleGAN2 \cite{karras2019analyzing} discriminator, without mini-batch statistic discrimination. We use $E$ for inverting images by having the last layer output a long flat vector, which we segment into blob parameters. In addition to reconstructing both real and synthetically generated images with LPIPS \cite{zhang2018unreasonable} and L2 penalties, we require the parameters $\hat{\boldsymbol \beta}$ to match the ground truth parameters $\boldsymbol \beta$ in the case of inverting generated images. Our overall loss is:
\begin{align}
    \mathcal{L}_\text{inversion} &= \mathcal{L}_\text{LPIPS}\left(\textbf{x}_\text{real}, G(E(\textbf{x}_\text{real}))\right) + \mathcal{L}_\text{LPIPS}(\textbf{x}_\text{fake}, G(E(\textbf{x}_\text{fake}))) \\ &+ \mathcal{L}_2(\textbf{x}_\text{real}, G(E(\textbf{x}_\text{real}))) + \mathcal{L}_2(\textbf{x}_\text{fake}, G(E(\textbf{x}_\text{fake}))) \nonumber \\ &+ \lambda \mathcal{L}_2(\boldsymbol\beta_\text{fake}, E(\textbf{x}_\text{fake})), \nonumber
\end{align}
with $\lambda = 10$ controlling the strength of the blob reconstruction loss. Taking the L2 loss on blob parameters as a flattened vector would heavily emphasize reconstructing the high-dimensional features, over the important, low-dimensional, scalar quantities of blob locations and sizes. Instead, we compute L2 separately over each blob attribute and take the mean.

We then further optimize the blob parameters to reconstruct the target image, with LPIPS and L2 losses and the Adam optimizer \cite{adam} with learning rate $0.01$ for $200$ steps. While better fitting the input image, this method potentially deviates from the manifold of latents that yield realistic images \cite{abdal2019image2stylegan,roich2021pivotal}, thus severely impeding editing abilities. Previously proposed solutions offer regularizations to keep the latents on this ``manifold''~\cite{wulff2020improving,zhu2020domain,xia2021gan}. However, we find our blob representation to be more robust in this sense, and latents yielded by this na\"{i}ve optimization still amenable to editing.

\section{Blob parametrization}
\label{app:blobparam}
We represent the blob aspect ratio $a$ as two scalar outputs $a_0, a_1$, sigmoided and then normalized to have a fixed product $a_0 a_1$; we find this to train more stably than one aspect ratio. We represent the blob angle $\theta$ with two scalars $e_0, e_1$, from which we construct a unit-normalized axis of rotation $\mathbf{e}$. We find this representation to train far more stably than others, such as regressing to a scalar $\theta$ or other parametrizations of $\Sigma$ like log-Cholesky \cite{mejjati2021gaussigan,shelhamer2019blurring}.

We also experimented with alternate representations, such as closed-form ellipses and rectangles as well as Gaussian mixture models. However, we found gradient flow to blob parameters ill-behaved with rectangles and other explicitly defined shapes, even with tricks like gradual opacity falloff, and these models failed to train. With GMMs, depth ordering and occlusions are lost, and blob size and shape depend on other blobs, harming performance. Our model is robust w.r.t.\ $c$, and $0.005 \leq c \leq 0.05$ all train well.

\subsection{Limitations}

Though our blob representation allows for powerful unsupervised, disentangled scene representations, our model still suffers from various shortcomings. For example, trained networks struggle to disentangle smaller objects ({\em e.g.}\ lamps on desks), perspective from object shape, and, occasionally, foreground appearance from background. Further, as shown in the main paper, blobs display a predilection toward certain canvas regions, though whether this is an artifact of dataset bias or model design remains unclear.

\section{Comparison to previous work}
\label{app:furtherresults}
\label{app:styleswap}

\begin{figure}[h]
\centering

\includegraphics[trim={0 20cm 0 0},clip,width=\textwidth]{figs/localstyleswap.pdf}
\includegraphics[trim={0 0 0 11.5cm},clip,width=\textwidth]{figs/localstyleswap.pdf}
\caption{\textbf{Swapping blob styles:} Interchanging $\psi_i$ vectors without modifying layout leads to localized edits which change the appearance of individual objects in the scene.}
\label{fig:localswap_row2}
\end{figure}

In Figures \ref{fig:sganswap} and \ref{fig:blobswap}, we show random samples of untruncated images before and after style swapping. At a given level of photorealism as measured by FID, our model is able to produce layouts far more consistent with the original image thanks to its disentangled, compositional representation. 

\begin{figure}[h]
\centering
\includegraphics[width=0.7\textwidth]{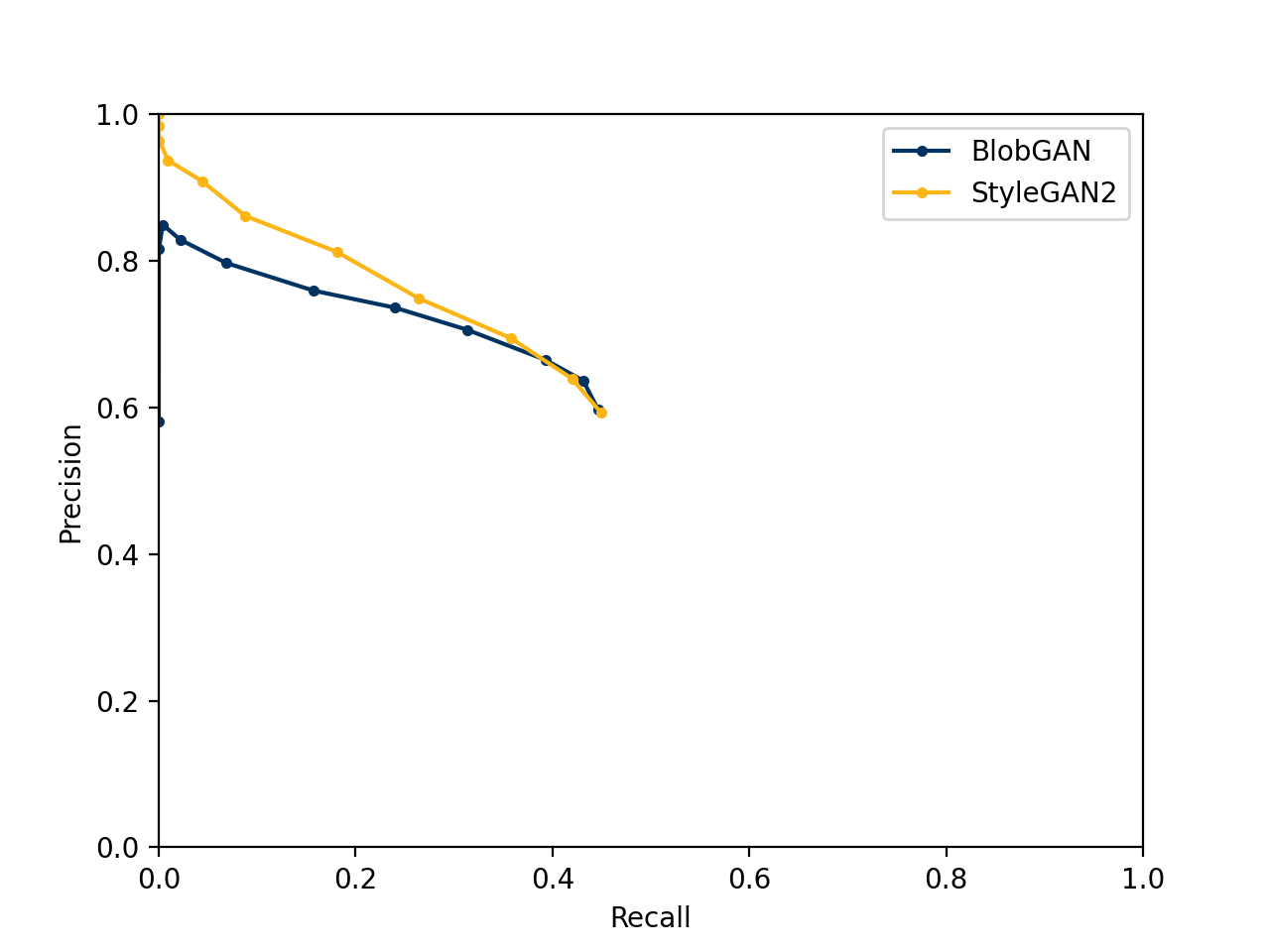}
\caption{We plot the precision-recall curve, by varying truncation values $w$, on LSUN bedrooms. Our untruncated model outperforms StyleGAN2 \cite{karras2019analyzing} with truncation values $w < 0.7$ in both precision and recall of generated images. While still outperforming StyleGAN2 on FID (Table \ref{tab:fid}), our model operates at a different point on this curve than StyleGAN2 -- higher precision and lower recall -- supporting the hypothesis that BlobGAN's FID suffers due to its inability to model long-tail, oddly-formed scenes.}
\label{fig:pr_curve}
\end{figure}

Lastly, we visualize the trade-off between the precision and recall metric~\cite{kynkaanniemi2019improved} as we change the truncation value in Figure~\ref{fig:pr_curve}. Our model generates more perceptually realistic images than StyleGAN at all truncation values $0.0 \leq w \leq 1.0$, although the maximal recall at $w = 1.0$ is lower. In particular, our untruncated model performs better at both precision and recall than {\em all StyleGAN-generated images with $w < 0.7$}. These results provide evidence for the suggestion that our model's FID is higher because it cannot properly model outlier bedroom scenes using the blob representation.

\section{Model implementation}
\label{app:modelimpl}
\subsection{Hyperparameters and training} For the bedroom model trained in the paper, we use $d_\text{in} = 768$ and $d_\text{style} = 512$. Our generator with $k=10$ blobs has 57.2 million parameters: 21.3 million in $F$ and the remaining 35.9 million in $G$.

The model trained on LSUN conference rooms uses $k=20$ and has 34.5M parameters in $F$; all other hyperparameters are as in the bedroom model.

The model trained on the union of LSUN kitchens, living rooms, and dining rooms uses $k=45$ due to the increased complexity of the combined dataset, and thus reduces $d_\text{in} = 256$ and $d_\text{style} = 256$. This model has 61.3 million parameters in the generator: 31.3M in $F$ and 30.0M in $G$.

We train all models for 1.5 million gradient steps with batch size 24 per-GPU across 8 NVIDIA A100 GPUs, except the bedrooms models (both BlobGAN and StyleGAN2), which are trained for 2.8 million steps. On the bedrooms model, we experiment with $k=20$ blobs as well as $k=10$ blobs with no jitter. We find that results are less interpretable with 20 blobs and disentanglement is lower, perhaps since the model can ``approximate'' slightly higher-frequency data by using more blobs. This model also has a worse FID of 3.73. We also train a model with $k=10$ blobs and no jitter, which attains comparable FID to the model with jitter, but with slightly reduced editing capabilities. Across all experiments, we find that changing $k$ minorly impacts FID. Extra blobs mostly go unused, but too few blobs mean objects cannot be properly separated.

\subsection{Image sampling} We sample all images shown in the paper and Supplementary Material with truncation. We truncate latents at the penultimate layer of $F$, since truncating in blob parameters space leads to undesirable behavior ({\em e.g.} biasing blob coordinates toward the center of the image). Then, truncation of random noise vector $z$'s output of blob parameters $\boldsymbol \beta$ with a weight of $w$ gives:
\begin{equation}
    \boldsymbol \beta_\text{trunc} = F_{L}\left((1-w)\mathop{\mathbb{E}}_{z^\prime \sim \mathcal{N}(0,I)}[F_{0:-1}(z^\prime)] + w F_{0:L-1}(z)\right)
\end{equation}
Where $F_{l:m}$ represents layers $l$ through $m$, inclusive, of the network which has $L$ layers total. In practice, we approximate the expectation by sampling 100,000 random noise vectors. We use $w = 0.6$ or $w = 0.7$ for all bedroom images. $w = 0.5$ for images of conference rooms, and $w=0.4$ for other indoor scenes,  except when indicated otherwise ($w=1$ means no truncation).

\subsection{Object style swapping} When swapping styles between objects, rather than splatting the target (new) object's style $\psi_{i, \text{tgt}}$ directly onto the source (original) image's background style $\psi_\text{bg, src}$, we interpolate first between $\psi_{i, \text{tgt}}$ and then $\psi_\text{bg,tgt}$ ({\em i.e.}, the target image's background) at the border of the blob, and then splat this onto the background $\psi_\text{bg, src}$. 

We find this necessary since the model learns to treat features on the border of a blob, which are typically a convex combination of the blob feature and the background feature, as belonging to the blob; when an unanticipated background feature becomes part of the feature along the border, the model is more prone to producing artifacts. This simple procedure mitigates this undesirable behavior and is trivially fully automated.

\subsection{Spatial modulation} In StyleGAN2, convolution weights at layer $l$, $\theta_l \in \mathbb{R}^{d_l \times d_{l-1} \times k \times k}$, are multiplied by an affine-transformed style vector $w \in \mathbb{R}^d_l$ and then unit-normalized to perform modulation. Since our modulation varies spatially, we instead multiply input feature maps $x_{l-1} \in \mathbb{R}^{d_{l-1} \times h \times w}$ by a unit-normalized style grid $\Psi_l \in \mathbb{R}^{d_\text{style} \times h \times w}$ with a per-pixel affine transform, before convolving with unit-normalized weights $\theta_l$ to output new feature maps $x_l$. Affine transforms $f$ map from $d_\text{style}$ to $d_l$. More specifically, in StyleGAN2, modulated convolution is implemented as:
\begin{equation}
    x_l = x_{l-1} \ast \frac{f(w_l) \odot \theta_l }{\|f(w_l) \odot \theta_l \|_2}
\end{equation}
Since our styles are spatially varying, we cannot multiply convolution weights by the same broadcasted tensor throughout, and must modify our modulation:
\begin{equation}
x_l =\left(x_{l-1} \odot  \frac{f(\Psi_l)}{\|f(\Psi_l)\|_2}\right) \ast \frac{\theta_l}{\|\theta_l\|_2}
\end{equation}
We find this normalization scheme, also used in \cite{park2019semantic,park2020swapping}, to work well in practice despite not having the same statistical guarantees as the original derivation.

\subsection{Uncurated samples}

\begin{figure}[h]
\centering
\includegraphics[width=\textwidth]{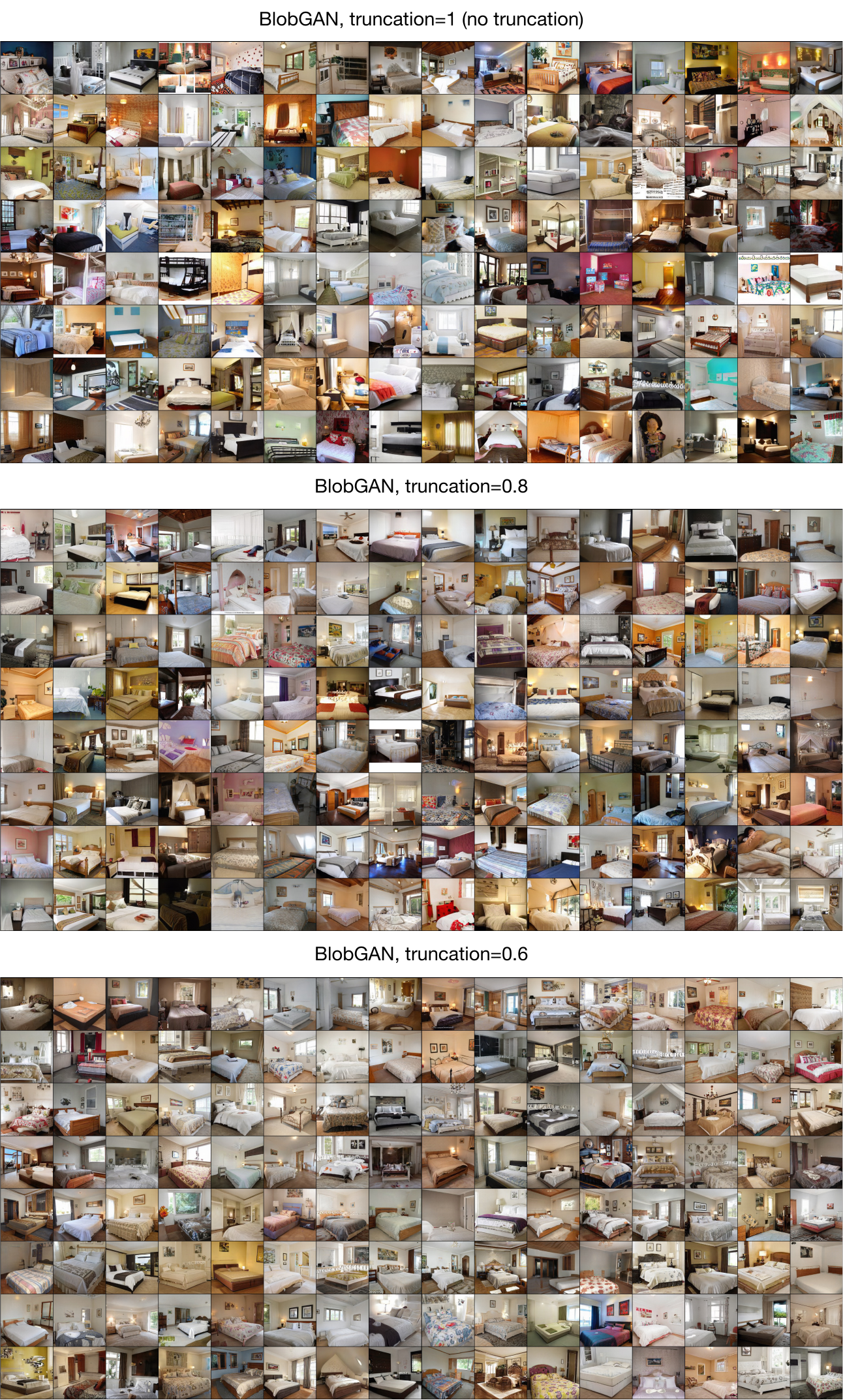}
\caption{We show uncurated random image samples from BlobGAN on LSUN bedrooms at various truncation levels. Please view zoomed in and in color for best results.}
\label{fig:uncuratedblobgan}
\end{figure}

\begin{figure}[h]
\centering
\includegraphics[width=\textwidth]{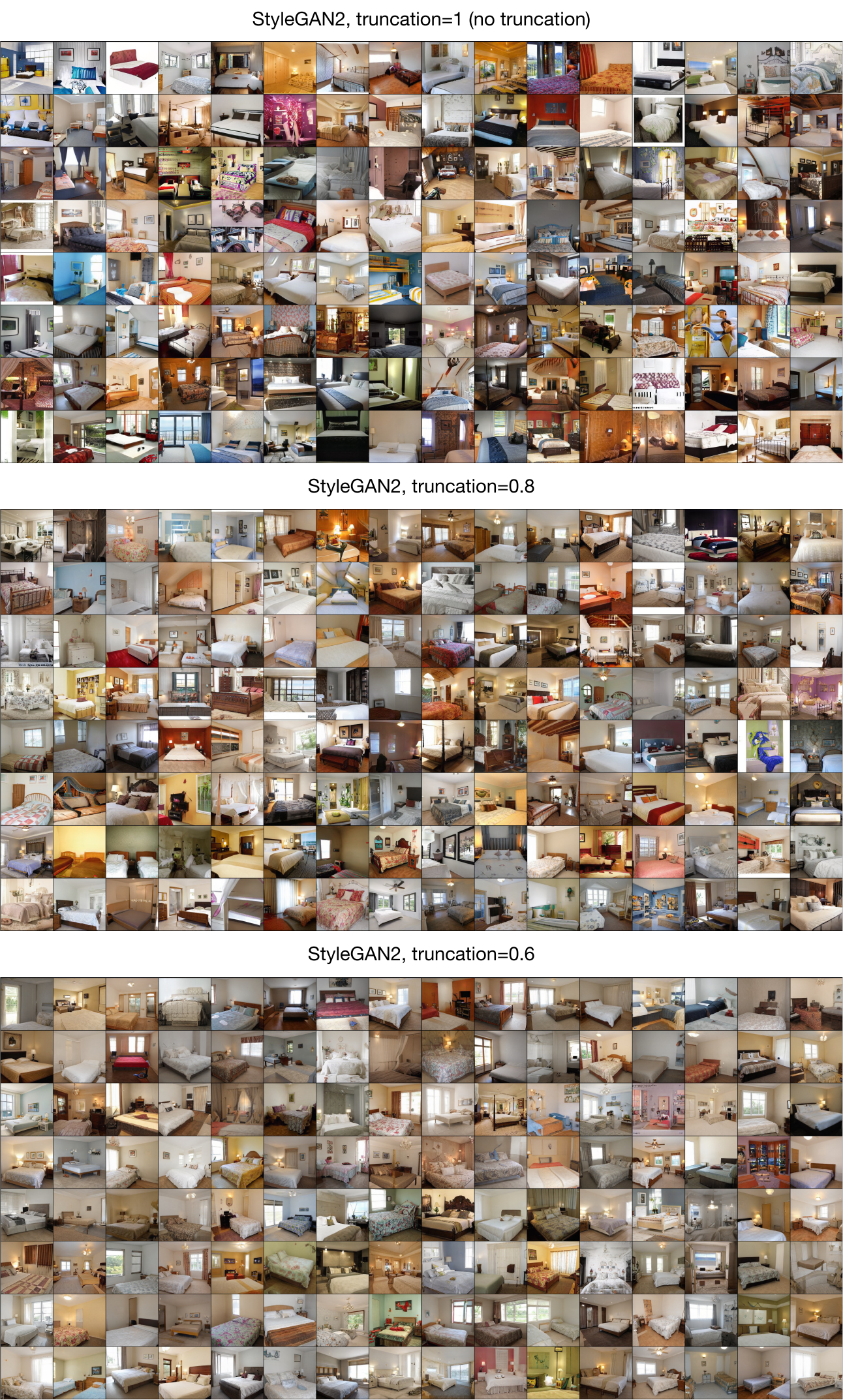}
\caption{We show uncurated random image samples from StyleGAN2 on LSUN bedrooms at various truncation levels. Please view zoomed in and in color for best results.}
\label{fig:uncuratedstylegan}
\end{figure}

\begin{figure}[h]
\centering
\includegraphics[width=\textwidth]{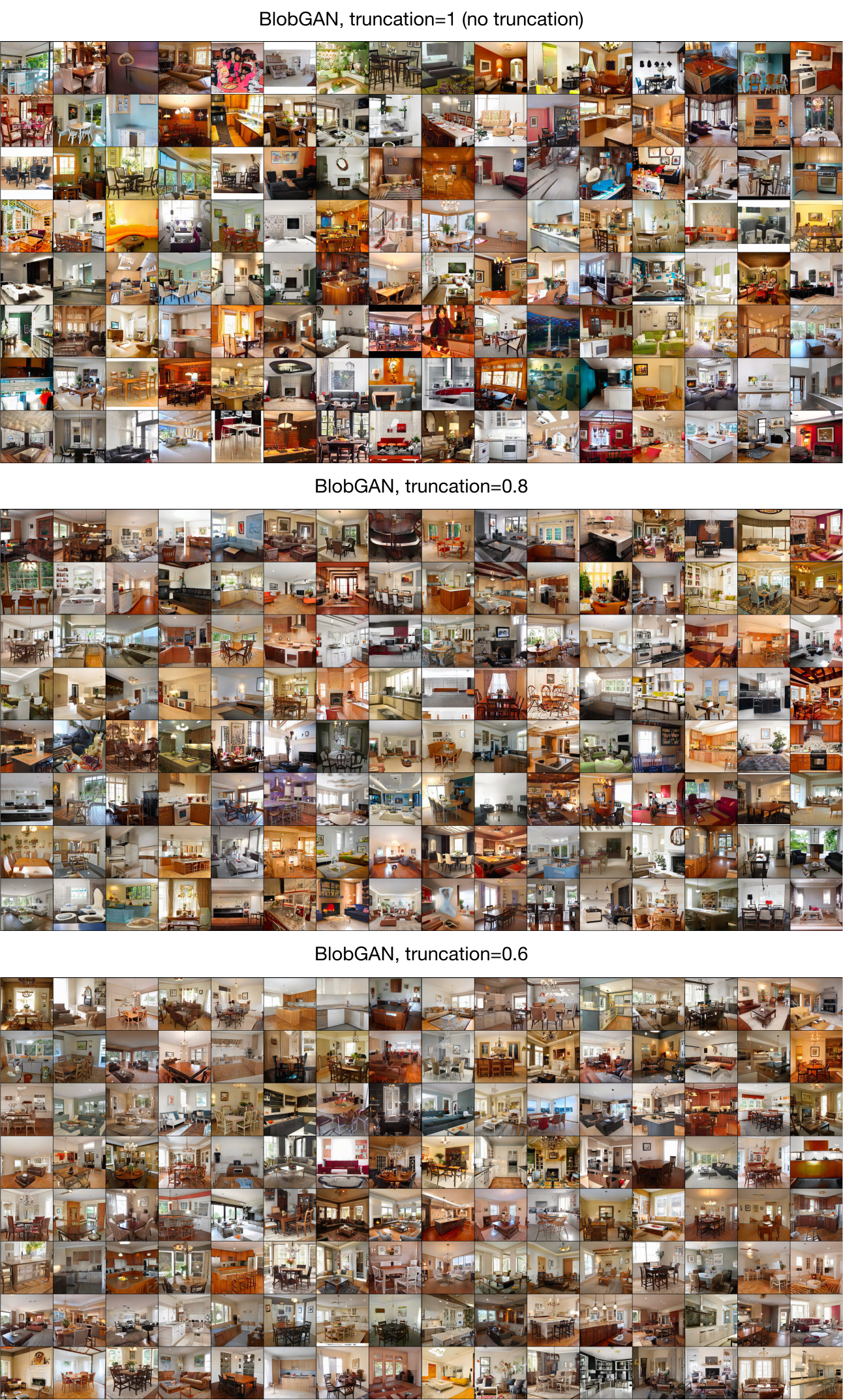}
\caption{We show uncurated random image samples from BlobGAN on LSUN kitchens, living rooms, and dining rooms at various truncation levels. Please view zoomed in and in color for best results.}
\label{fig:uncuratedblobgankld}
\end{figure}

\begin{figure}[h]
\centering
\includegraphics[width=\textwidth]{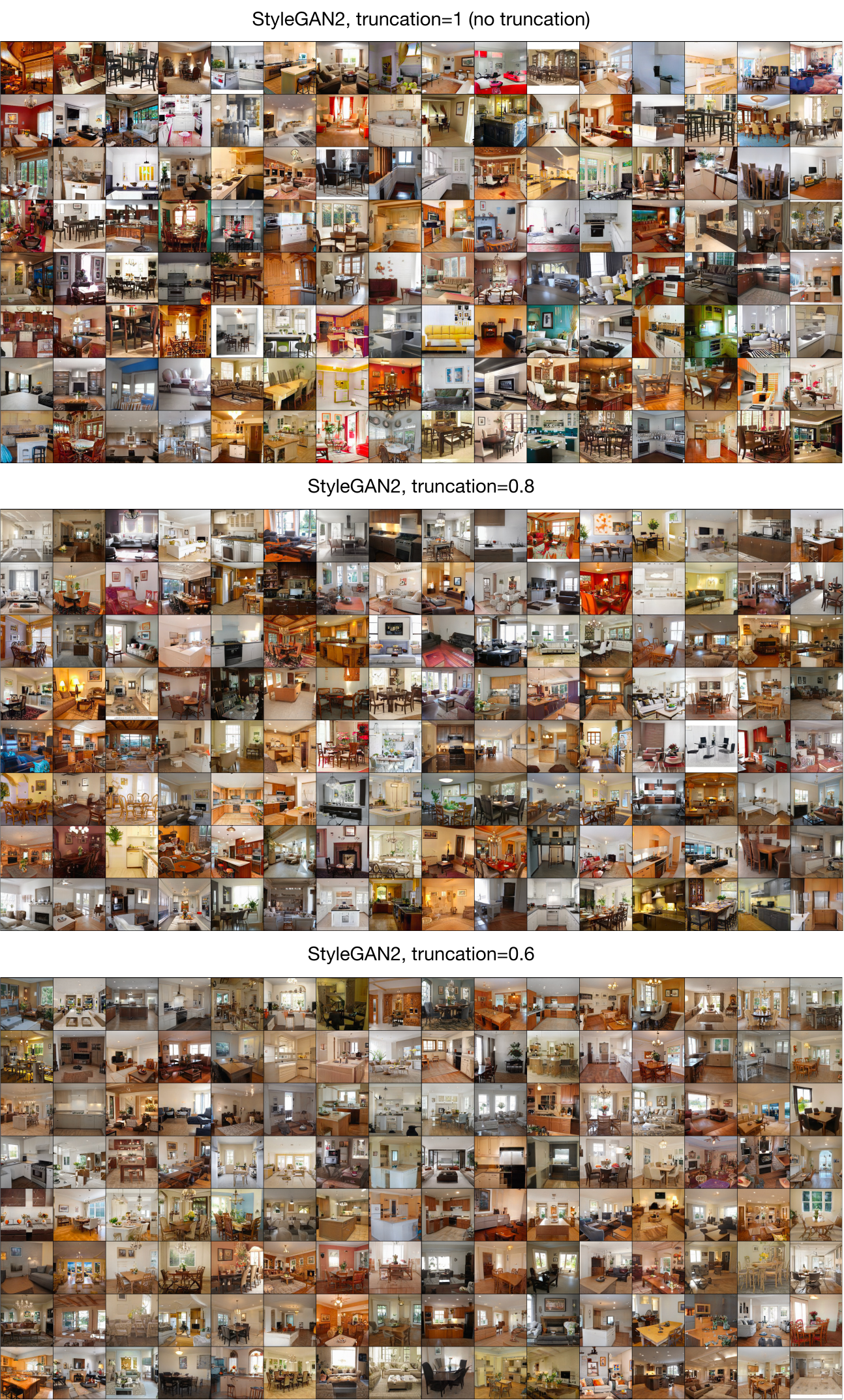}
\caption{We show uncurated random image samples from StyleGAN2 on LSUN kitchens, living rooms, and dining rooms at various truncation levels. Please view zoomed in and in color for best results.}
\label{fig:uncuratedstylegankld}
\end{figure}

\begin{figure}[h]
\centering
\includegraphics[width=\textwidth]{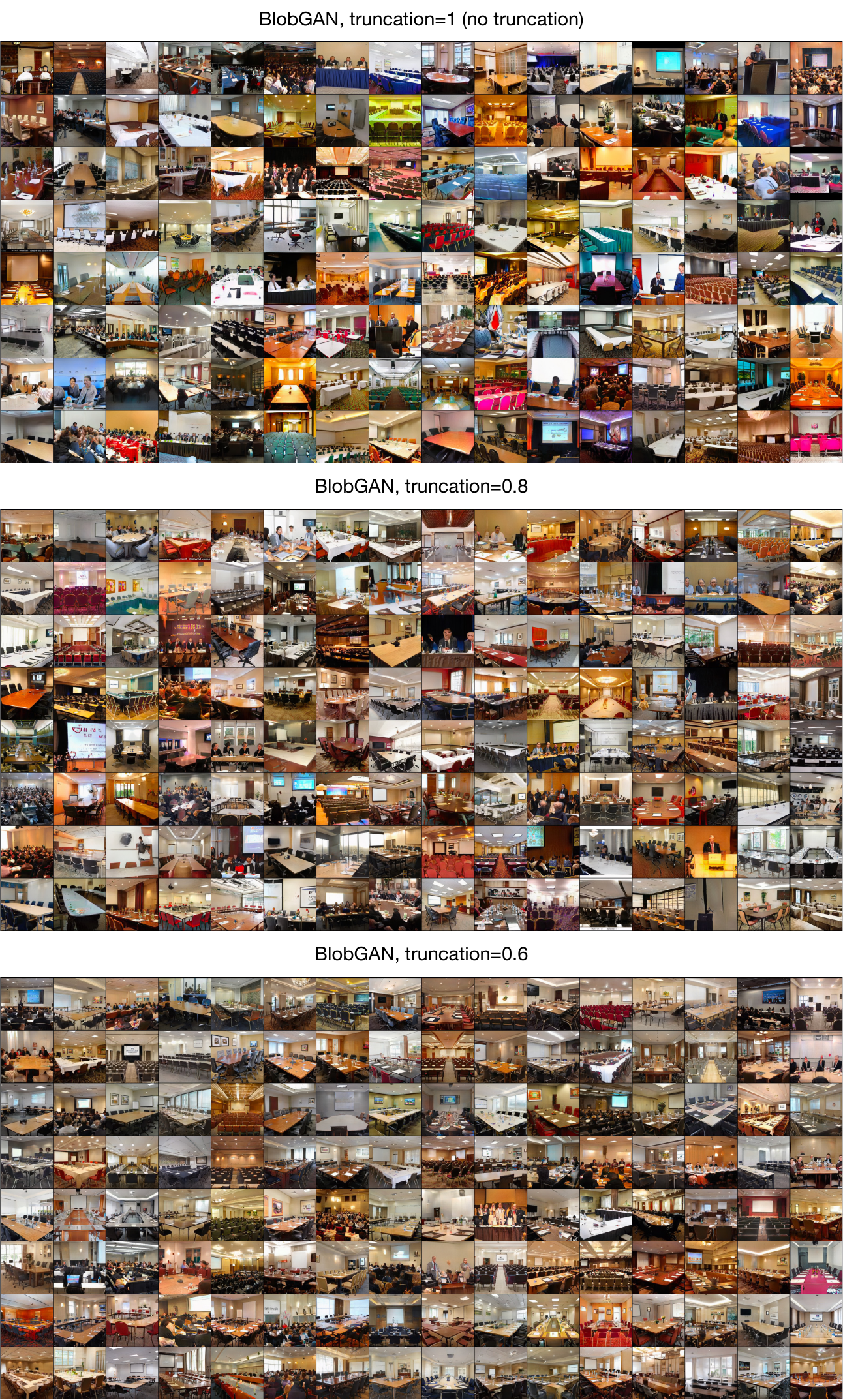}
\caption{We show uncurated random image samples from BlobGAN on LSUN conference rooms at various truncation levels. Please view zoomed in and in color for best results.}
\label{fig:uncuratedblobganconf}
\end{figure}

\begin{figure}[h]
\centering
\includegraphics[width=\textwidth]{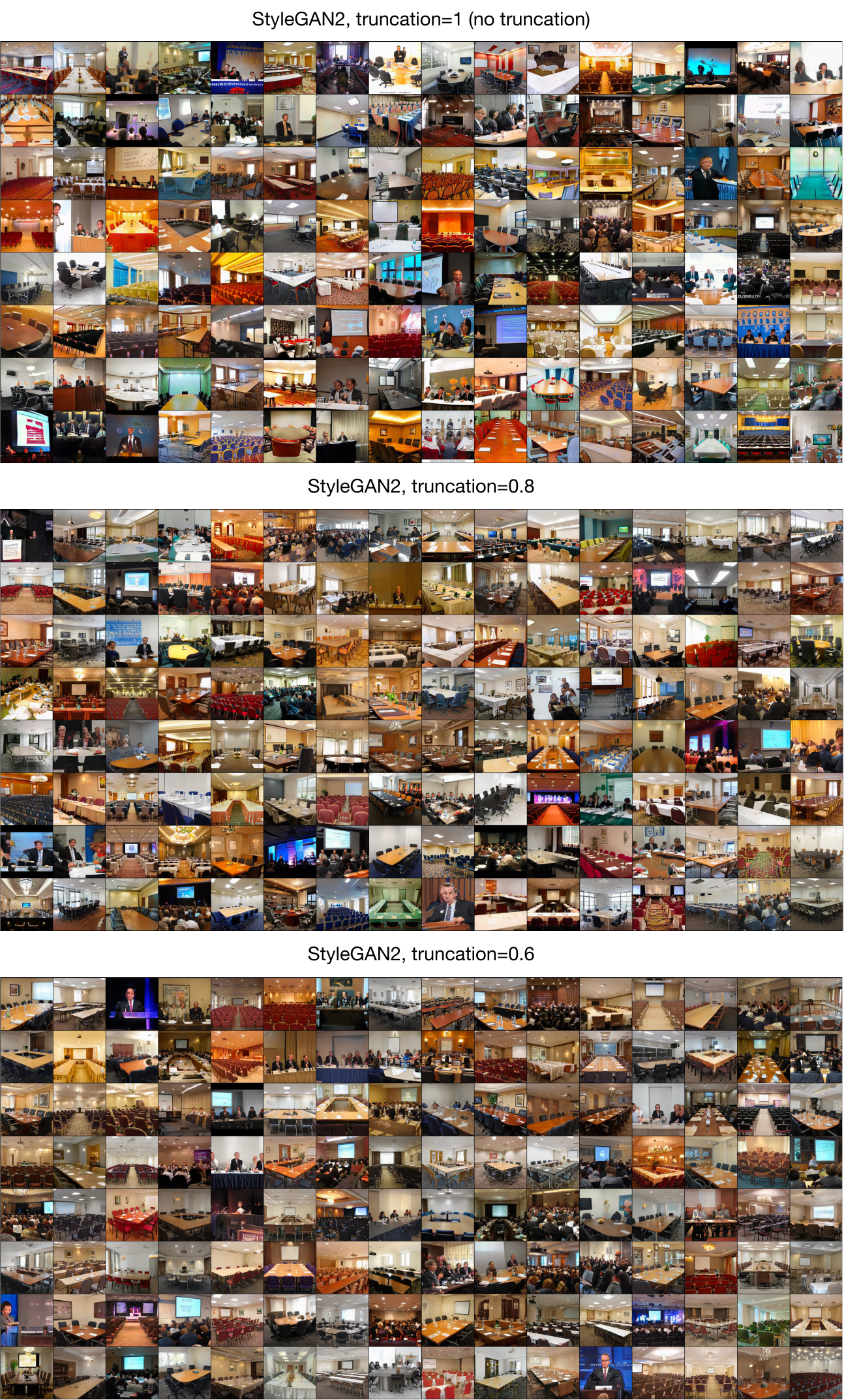}
\caption{We show uncurated random image samples from StyleGAN2 on LSUN conference rooms at various truncation levels. Please view zoomed in and in color for best results.}
\label{fig:uncuratedstyleganconf}
\end{figure}

In Figures \ref{fig:uncuratedblobgan} and \ref{fig:uncuratedstylegan}, we show randomly sampled images from our model and StyleGAN2 trained on LSUN Bedrooms. We show the same on LSUN kitchens, living rooms, dining rooms, and conference rooms in Figures \ref{fig:uncuratedblobgankld}, \ref{fig:uncuratedstylegankld}, \ref{fig:uncuratedblobganconf}, and \ref{fig:uncuratedstyleganconf}.

\end{document}